\documentclass[journal]{IEEEtran}
\usepackage{amsmath}
\usepackage{bm}
\usepackage{amssymb}
\usepackage{graphicx,color}
\usepackage{soul}
\usepackage{algorithmic,algorithm}

\usepackage{cite}
\usepackage{balance}
\usepackage{subfig}
\usepackage{url}
\begin{document}
\title{Generalized Gaussian Kernel Adaptive Filtering}
%\author{Tomoya Wada, Kosuke Fukumori, Toshihisa Tanaka,~\IEEEmembership{Senior Member,~IEEE}, and Simone Fiori% <-this % stops a space
\author{Tomoya Wada, Kosuke Fukumori, Toshihisa Tanaka, and Simone Fiori% <-this % stops a space
\thanks{
This work is supported by JSPS KAKENHI Grant Number 17H01760
and National Center for Theoretical Sciences (NCTS), Taiwan, through a  2016 ``Research in Pairs'' program.
T.~Wada, K.~Fukumori, and T.~Tanaka are with the Department of Electrical and Electronic Engineering, Tokyo University of Agriculture and Technology, 2–24–16 Nakacho, Koganei-shi, Tokyo, 184-8588, Japan (e-mail: \{wada15, fukumori17\}@sip.tuat.ac.jp, tanakat@cc.tuat.ac.jp).
S.~Fiori is with Universit\`{a} Politecnica delle Marche, Via Brecce Bianche, 60131 Ancona, Italy (e-mail: s.fiori@univpm.it).}% <-this % stops a space
}
%\markboth{IEEE Transactions on Signal Processing}%
%{Wada, Fukumori, Tanaka, Fiori: Generalized Gaussian Kernel Adaptive Filtering}
\maketitle
\begin{abstract}
The present paper proposes generalized Gaussian kernel adaptive filtering, where the kernel parameters are adaptive and data-driven.
The Gaussian kernel is parametrized by a center vector and a symmetric positive definite (SPD) precision matrix, which is regarded as a generalization of the scalar width parameter.
These parameters are adaptively updated on the basis of a proposed least-square-type rule to minimize the estimation error. The main contribution of this paper is to establish update rules for precision matrices on the SPD manifold in order to keep their symmetric positive-definiteness.
Different from conventional kernel adaptive filters, the proposed regressor is a superposition of Gaussian kernels with all different parameters, which makes such regressor more flexible.
The kernel adaptive filtering algorithm is established together with a $\ell_1$-regularized least squares to avoid overfitting and the increase of dimensionality of the dictionary.
Experimental results confirm the validity of the proposed method.
\end{abstract}
\section{Introduction}
\label{sec:intro}
\IEEEPARstart{An}{adaptive} filter or adaptive filtering is a system or technique that updates its parameters at every time step to approximate a static or dynamic unknown system~\cite{haykin2002}.
Although in traditional adaptive filters a linear model is assumed,
many situations in the real environments require nonlinear adaptive filters.
Several types of nonlinear adaptive filters have been reported.
Among them, kernel adaptive filtering developed in a reproducing kernel Hilbert space (RKHS) is known as an efficient online nonlinear approximation approach~\cite{kivinen2004, liu2010}.

In kernel adaptive filtering, the model is represented by the superposition of the kernels corresponding to the observed signals (or samples),
where the adaptive algorithm is intended to estimate coupling coefficients of kernels.
Typical kernel adaptive filtering algorithms include the kernel least mean square (KLMS)~\cite{liu2008, bouboulis2011,chen2012Q,tobar2014},
the kernel normalized least mean square (KNLMS),
the kernel affine projection algorithms (KAPA)~\cite{liu2008kernel,cacho2012},
and the kernel recursive least squares (KRLS)~\cite{engel2004}.
The main bottleneck of the kernel adaptive filtering algorithms is their linearly growing structure with each new input signal, which poses computational issues and may cause overfitting.
A straightforward -- yet practical -- approach to cope with this problem is to limit the number of observed signals. This set of observed signals is called a \textit{dictionary}.
Typical criteria for the dictionary learning include the novelty criterion~\cite{platt1991}, the approximate linear dependency (ALD) criterion~\cite{engel2004}, the surprise criterion~\cite{liu2009extended}, and the coherence-based criterion~\cite{richard2009}.
These criteria accept only the novel and informative input signals as dictionary members.
Another approach is the $\ell_1$-regularization~\cite{gao2013, gao2014}.
In this approach, the filter coefficients are regularized by the $\ell_1$-norm,
which set some coefficients to zero, and then the corresponding entries in the dictionary are discarded. Therefore, the model dynamically changes and new members may be added to a dictionary as well as old members may be suppressed from a dictionary.

Another feature of kernel adaptive filtering is the ability to update the parameters of each kernel to decrease the estimation error of the output.
In standard kernel adaptive filtering, the center vector of each kernel is given as an observed signal.
Some related works proposed to adaptively move all the center vectors in the dictionary to minimize the square error~\cite{saide2012,saide2013,ishida2015}.
It is known that the kernel width is an important parameter to govern the performance of kernel machines~\cite{benoudjit2003kernel, ghosh2008kernel, chen2016kernel,fan2016, wada2017}.
Some attempts to adaptively estimate the kernel width has been reported~\cite{fan2016, wada2017}.
Moreover, in a recent work~\cite{wada2018}, Wada \emph{et al.}\ have proposed an adaptive update method for both the Gaussian center and width.
%Although the above methods are effective for increasing the performance of kernel adaptive filters,
Most Gaussian kernel machines present the following form:
\begin{equation}
\kappa(\cdot, \bm{c}; \zeta) = \exp\left(-\zeta\|\cdot- \bm{c} \|^2\right),\label{eq:gaussian}
\end{equation}
where $\bm{c}$ and $\zeta$ are parameters called the \textit{center} and the \textit{width} of the Gaussian kernel, respectively.
It should be noted that this form implicitly assumes uncorrelatedness between components in the sample vector.
In other words, the kernel presents only two parameters, namely, mean and variance (precision).
However, observed samples usually present some sort of mutual correlation.

In this paper, we employ a generalized Gaussian kernel defined as
\begin{equation}
\kappa(\cdot, \bm{c}; \bm{Z}) = \exp\left(-(\cdot- \bm{c})^{\top}\bm{Z}(\cdot- \bm{c})\right),\label{eq:generalized_gaussian}
\end{equation}
where $\bm{Z}\in\mathbb{R}^{L\times L}$ is the inverse covariance matrix, which is a symmetric positive definite (SPD) matrix.
Here, we refer to $\bm{Z}$ as a precision matrix.
Unlike \eqref{eq:gaussian}, this form has more degrees of freedom, and therefore it is more flexible in modeling signals.
We will establish a dictionary learning method for generalized Gaussian kernel adaptive filtering.
In a dictionary for the proposed kernel adaptive filtering, each entry consists of a pair formed by a center vector and a precision matrix.
For each input signal, all entries in the dictionary are updated to minimize the estimation error through least-square-type rules.
The main contribution of the proposed method is a model of the filter consisting of kernels with all different precision matrices, an update rule for center vectors as well as that for precision matrices formulated on the Lie group of symmetric positive-definite matrices.
This double adaptation strategy for the center vectors and the precision matrices in the proposed model is merged with a $\ell_1$-regularized least squares technique for updating the filter coefficients, which allows one to avoid overfitting and the excessive increasing of dimensionality of the dictionary.

The paper is organized as follows: Section \ref{sec:KAF} presents general concepts in kernel adaptive filtering.
Section \ref{sec:proposed} proposes a dictionary learning method for the generalized Gaussian kernel adaptive filtering.
The main contribution of the proposed method, which is the update rule for precision matrices of kernels, is presented in Subsection \ref{update_matrix_width}.
Section \ref{sec:example} shows numerical examples to support the efficacy of the proposed methods.
Section \ref{sec:conclusion} concludes the paper.
\section{Kernel Adaptive Filters}\label{sec:KAF}
A kernel adaptive filter is a kind of nonlinear filter that exploits a kernel method,
which is a technique to construct effective nonlinear systems based on a
RKHS induced from a positive definite kernel~\cite{aronszajn1950}.
In recent years, the efficiency of kernel adaptive filters has become known
since kernel adaptive filters have the following features~\cite{liu2010}:
\begin{itemize}
	\item They are universal approximators;
	\item They present no local minima;
	\item They present moderate complexity in terms of computation and memory.
\end{itemize}
In this section, we first discuss signal modeling in the context of kernel adaptive filtering and
next we briefly review well-known kernel adaptive algorithms.

\subsection{Nonlinear Filtering Model in Kernel Adaptive Filters}
Let $\mathcal{U} \subset \mathbb{R}^L$, $\bm{u}^{(n)}\in\mathcal{U}$, and $d^{(n)}\in\mathbb{R}$ denote the input space, an input signal, and the corresponding desired output signal at the time-instant $n$, respectively.

In kernel adaptive filtering, an input $\bm{u}^{(n)}$ is mapped to a RKHS $(\mathcal{H}, \langle\cdot, \cdot\rangle)$ on $\mathcal{U}$ induced from a positive definite kernel $\kappa(\cdot,\cdot):\mathcal{U}\times\mathcal{U}\to\mathbb{R}$ as
a high dimensional feature space to treat the nonlinearity of $\bm{u}^{(n)}$.
Here,  $\langle\cdot, \cdot\rangle$ denotes the inner product in the RKHS.
The output of the system is modeled as the inner product of a filter $\Omega^{(n)}\in\mathcal{H}$ with a nonlinear mapping of an input signal $\phi(\bm{u}^{(n)})\in\mathcal{H}$ as
\begin{equation}
 f(\bm{u}^{(n)})=\langle\phi(\bm{u}^{(n)}),\Omega^{(n)}\rangle.
\end{equation}
In general, the inner product in a high dimensional space is not given as an explicit form.
Rather, the inner product in RKHS can be calculated by using the import properties of RKHS, namely:
(i) all elements in a RKHS are constructed by a kernel $\kappa(\cdot,\bm{u})$,
(ii) $\phi(\bm{u}) = \kappa(\cdot, \bm{u})$,
(iii) $\langle\kappa(\cdot,\bm{u}_{i}), \kappa(\cdot, \bm{u}_{j})\rangle=\kappa(\bm{u}_{i},\bm{u}_{j})$~\cite{aronszajn1950,richard2009}.

\begin{figure}
	\centering
	\includegraphics[width=\columnwidth,clip]{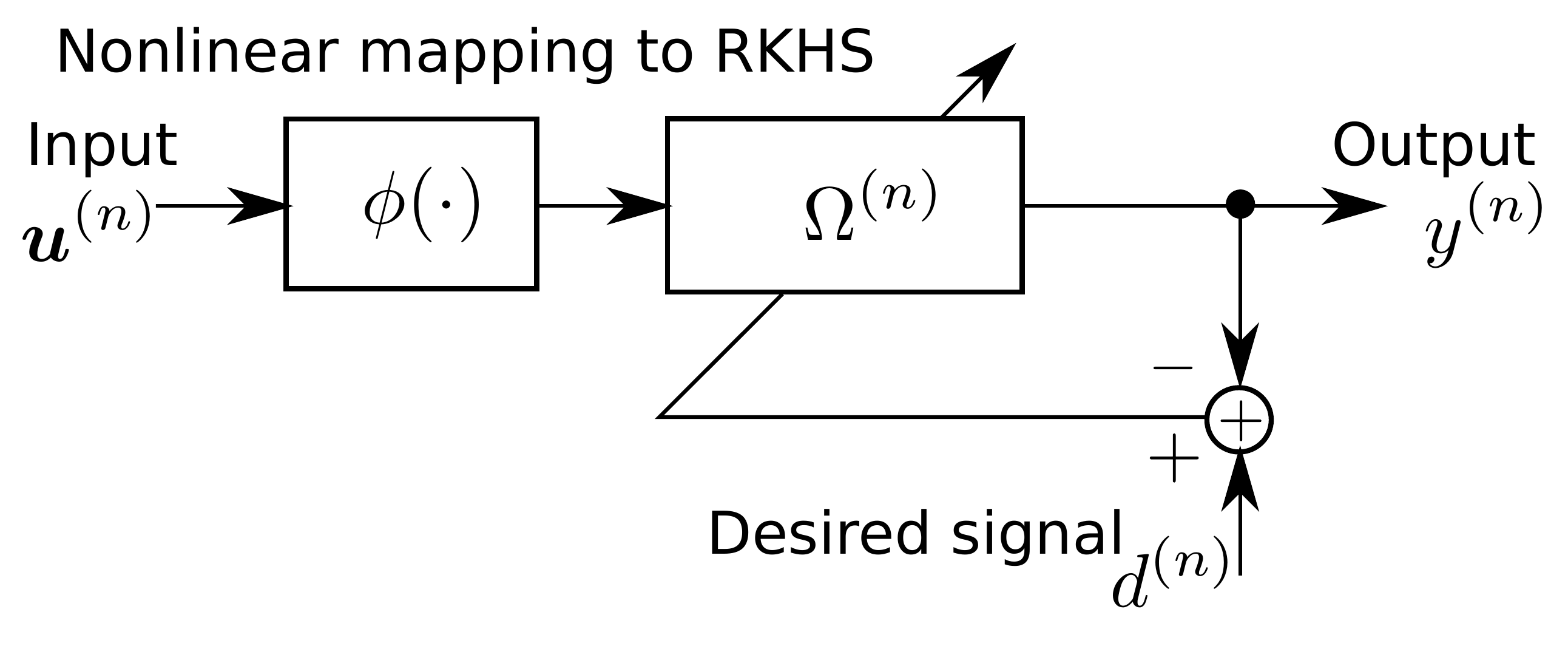}
	%\vspace{-2pt}
	\caption{Conceptual diagram of kernel adaptive filters.}
	\label{fig:gainen}
	%\vspace{-3pt}
\end{figure}

We consider the problem of adaptively estimating a filter $\Omega^{(n)}$.
The Figure~\ref{fig:gainen} shows a conceptual diagram of the kernel adaptive filter.
By the representer theorem~\cite{richard2009}, $\Omega^{(n)}$ can be written as
\begin{equation}
 \Omega^{(n)} = \underset{j\in\mathcal{J}^{(n)}}{\sum}{h^{(n)}_{j}}\kappa(\cdot,\bm{c}_{j}),
\label{filter}
\end{equation}
where the $h^{(n)}_{j}\in\mathbb{R}$ are scalar weight coefficient for $\kappa(\cdot,\bm{c}_{j})$.
From this, it is seen that estimating $\Omega^{(n)}$ is essentially equivalent to estimating a set of coefficients $h^{(n)}_{j}$.
Here, $\mathcal{D}^{(n)}=\{\bm{c}_j\}_{j\in\mathcal{J}^{(n)}}$ is a set of input signals accepted only if they satisfy a predefined criterion. This set is called \emph{dictionary}.
The index set of dictionary elements and the dictionary size at time $n$ are defined as
$\mathcal{J}^{(n)}:=\{j_{1}^{(n)},j_{2}^{(n)},\ldots,j_{r^{(n)}}^{(n)}\}\subset\{0,1,\ldots,n-1\}$ and $r^{(n)}=|\mathcal{J}^{(n)}|$, respectively.
The filter output is represented as
\begin{eqnarray}
\nonumber
y^{(n)}&=&\langle\phi(\bm{u}^{(n)}),\Omega^{(n)}\rangle=\underset{j\in\mathcal{J}^{(n)}}{\sum}{h^{(n)}_{j}}\kappa(\bm{u}^{(n)},\bm{c}_{j})\\
       &=&{\bm{h}^{(n)}}^\top\bm{\kappa}^{(n)},
\end{eqnarray}
where
\begin{align}
\bm{h}^{(n)} &:=[h_{j_{1}^{(n)}}^{(n)},h_{j_{2}^{(n)}}^{(n)},\ldots,h_{j_{r^{(n)}}^{(n)}}^{(n)}]^\top, \\
\bm{\kappa}^{(n)} &:=[\kappa(\bm{u}^{(n)},\bm{c}_{j_{1}^{(n)}}),\kappa(\bm{u}^{(n)},\bm{c}_{j_{2}^{(n)}}),\ldots,\kappa(\bm{u}^{(n)},\bm{c}_{j_{r^{(n)}}^{(n)}})]^\top,
\end{align}
and both vectors $\bm{h}^{(n)},\bm{\kappa}^{(n)}$ belong to $\mathbb{R}^{r^{(n)}}$.
\subsection{$\ell_1$-regularized KNLMS (KNLMS-$\ell_1$)}\label{subsec:KNLMS_L1}
The kernel adaptive filtering algorithms can only incorporate new elements into the dictionary.
This unfortunately means that it cannot discard obsolete kernel functions, within the context of a time-varying environment in particular.
Recently, to remedy this drawback, it has been proposed to construct a dictionary by $\ell_1$-regularization~\cite{gao2013,gao2014}.
In this scenario, a weighted $\ell_1$-norm is added to the cost function of KNLMS in order to effectively adapt nonstationary systems.
The cost function is written as follows:
\begin{equation}\label{costfun}
\Theta^{(n)}:=|d^{(n)}-{\bm{h}^{(n)}}^\top\bm{\kappa}^{(n)}|^2+\lambda \underbrace{\underset{j\in\mathcal{J}^{(n)}}{\sum}w_{j}^{(n)}|h_{j}^{(n)}|}_{:=\psi^{(n)}},
\end{equation}
where $\psi^{(n)}$ and $\lambda$ play the role of a weighted $\ell_1$ norm and of a regularization parameter, respectively.
Here, weights $\{w_{j}^{(n)}\}_{j\in\mathcal{J}^{(n)}}$ are dynamically adjusted as
$w_{j}^{(n)}={1}/{(|h_{j}^{(n)}|+\beta)}$~\cite{gao2014}, with a small constant $\beta$ to prevent the denominator from vanishing.
It is not possible to apply the stochastic gradient approach to minimize the cost function \eqref{costfun} since the weighted $\ell_1$ norm is nonsmooth.
However, since $\Theta^{(n)}$ is a convex function,
the forward-backward splitting~\cite{murakami2010} may be applied.
The update rule is then given as follows:
\begin{align}\label{eq:l1}
\bm{h}^{(n+1)}= {\rm prox }_{\mu\lambda\psi^{(n)}}\left[\overline{\bm{h}^{(n)}}+\frac{\mu\left(d^{(n)}- \overline{\bm{h}^{(n)}}^{\top}\overline{\bm{\kappa}^{(n)}}\right)\overline{\bm{\kappa}^{(n)}}}{\rho+\|\overline{\bm{\kappa}^{(n)}}\|^2}\right],
\end{align}
where
${\rm prox}_{\mu\lambda\psi^{(n)}}(\cdot)$ denotes the proximal operator~\cite{murakami2010} of $\lambda \psi^{(n)}$, $\overline{\bm{h}^{(n)}} :=[{\bm{h}^{(n)}}^\top,0]^\top$, $\overline{\bm{\kappa}^{(n)}}:=[{\bm{\kappa}^{(n)}}^\top, \kappa(\bm{u}^{(n)},\bm{u}^{(n)})]^\top$, the coefficient $\mu$ denotes a step size parameter, the coefficient $\rho$ denotes a stabilization parameter, and $\|\cdot\|$ denotes a standard vector 2-norm.
Concretely, assuming that a vector $\bm{\alpha} :=[\alpha_{1},\alpha_2,\ldots,\alpha_r]^\top\in\mathbb{R}^{r}$ is given, ${\rm prox}_{\mu\lambda\psi^{(n)}}(\bm{\alpha})$ can be expressed as
\begin{equation}
\left({\rm prox}_{\mu\lambda\psi^{(n)}}(\bm{\alpha})\right)_j={\rm sgn}\{\alpha_{j}\}\max\{|\alpha_{j}|-\mu\lambda w_{j}^{(n)},0\},
\end{equation}
where $(\cdot)_j$ denotes the $j$-th element of a vector.
The rule \eqref{eq:l1} promotes the sparsity of $h_{j}^{(n)}$, which results in
some coefficient $h_{j}^{(n)}$ approaching zero and the corresponding center vector
$\bm{c}_j$ getting removed from the dictionary.

\section{Model and Dictionary Learning for Generalized Gaussian Kernel Adaptive Filtering}\label{sec:proposed}
Most kernel machines using Gaussian kernel functions implicitly assume uncorrelatedness within the sample.
In other words, the kernel has only two parameters (namely, mean and variance) even though
observed samples usually present correlation.
In the following, a flexible model using a generalized Gaussian function given as in \eqref{eq:generalized_gaussian} is proposed.
Moreover, efficient algorithms for learning parameters are established.

\subsection{Model}
The proposed model is the superposition of generalized Gaussian kernels with time-varying $\bm{c}_j^{(n)}$ and $\bm{Z}_j^{(n)}$ given as
\begin{align}
y^{(n)}=&\underset{j\in\mathcal{J}^{(n)}}{\sum}{h_{j}}^{(n)}\kappa(\bm{u}^{(n)},\bm{c}^{(n)}_{j};\bm{Z}^{(n)}_j)\nonumber \\
=&\underset{j\in\mathcal{J}^{(n)}}{\sum}{h_{j}}^{(n)}\exp\left(-(\bm{u}^{(n)}- \bm{c}^{(n)}_j)^{\top}\bm{Z}_j^{(n)}(\bm{u}^{(n)}- \bm{c}^{(n)}_j)\right).
\label{eq:proposed_model}
\end{align}
The dictionary at time $n$ is a time-variable set of pairs, a center vector and a precision matrix for each kernel, which is described as
\begin{equation}
\mathcal{D}^{(n)}=\{(\bm{c}^{(n)}_{j_1},\bm{Z}^{(n)}_{j_1}),(\bm{c}^{(n)}_{j_2},\bm{Z}^{(n)}_{j_2}),\dots,(\bm{c}^{(n)}_{j_{r^{(n)}}},\bm{Z}^{(n)}_{j_{r^{(n)}}})\}.
\end{equation}
In the rest of this section, a dictionary learning method for generalized Gaussian kernel adaptive filtering is proposed. To adaptively compute the optimal parameters, we adopt the instantaneous square error as the loss function:
\begin{align}
\nonumber
&J^{(n)}(\mathcal{D}^{(n)}):=|e^{(n)}|^2=|d^{(n)}-y^{(n)}|^2=\\
&\left|d^{(n)}-\underset{j\in\mathcal{J}^{(n)}}{\sum}{h^{(n)}_{j}}\exp\left(-(\bm{u}^{(n)}- \bm{c}_j)^{\top}\bm{Z}^{(n)}_{j}(\bm{u}^{(n)}- \bm{c}_j)\right)\right|^2. \label{eq:loss_matrix}
\end{align}

{\bf Remark.}
We describe the sum space of RKHS~\cite{aronszajn1950} in order to discuss a space in which multikernel adaptive filters~\cite{yukawa2012,ishida2013}, including the proposed filter, exist.
We consider the case of sum space of two RKHS, for the sake of ease, without loss of generality.

Let $\mathcal{H}_1$ and $\mathcal{H}_2$ denote Hilbert spaces and let
$H:=\mathcal{H}_{1}\oplus\mathcal{H}_{2}$ denote their direct sum.
In this case, the norm of the direct sum of $f_{1}\in\mathcal{H}_{1}$ and $f_{2}\in\mathcal{H}_{2}$, $f=(f_{1},f_{2})\in{H}$, is represented as~\cite{aronszajn1950}:
\begin{equation}
\|f\|_{H}^2:=\|f_{1}\|_{\mathcal{H}_{1}}^2+\|f_{2}\|_{\mathcal{H}_{2}}^2.
\label{norm}
\end{equation}
In particular, if $\mathcal{H}_{1}\cap\mathcal{H}_{2}=\{0\}$, the sum space, ${\mathcal{H}}:=\{f=f_{1}+f_{2}\mid{f_{1}\in\mathcal{H}_{1},f_{2}\in\mathcal{H}_{2}}\}$,
is isomorphic to the direct space, $H$~\cite{aronszajn1950}.
Consequently, the norm in ${\mathcal{H}}$ is represented as
\begin{equation}
\|f\|_{{\mathcal{H}}}^2:=\|f_{1}\|_{\mathcal{H}_{1}}^2+\|f_{2}\|_{\mathcal{H}_{2}}^2.
\end{equation}
Also, let a kernel in $\mathcal{H}_{1}$ and a kernel in $\mathcal{H}_{2}$ be denoted as $\kappa_{1}$ and $\kappa_{2}$, respectively.
The value of any $f\in{\mathcal{H}}$ can be evaluated by the kernel $\kappa=\kappa_{1}+\kappa_{2}$~\cite{aronszajn1950}:
\begin{align}
f(\bm{u})=\langle{f},{\kappa(\cdot,\bm{u})}\rangle_{{\mathcal{H}}}=\langle{f_{1}},{\kappa_{1}(\cdot,\bm{u})}\rangle_{\mathcal{H}_{1}}
+\langle{f_{2}},{\kappa_{2}(\cdot,\bm{u})}\rangle_{\mathcal{H}_{2}}.
\label{saisei}
\end{align}
Assume that $M$ different kernels, $\{\kappa_m(\cdot,\cdot)\}_{m=1}^{M}$, are given.
Also, let $\mathcal{H}_{m}$ and $\mathcal{H}$ denote a RKHS determined by the $m$-th kernel and the corresponding sum space, respectively.
In this case, from (\ref{saisei}), the output is represented
by the filter $\Omega\in\mathcal{H}$ and by a nonlinear mapping of input $\phi(\bm{u}^{(n)})=\kappa(\cdot,\bm{u}^{(n)})\in\mathcal{H}$ as
\begin{align}
y^{(n)}&=\langle \Omega, \kappa(\cdot,\bm{u}^{(n)})\rangle_{\mathcal{H}}=
       {\sum}_{m=1}^{M}\langle \Omega_m, \kappa_m(\cdot,\bm{u}^{(n)})\rangle_{\mathcal{H}_m},
\end{align}
where $\Omega_m$ is constructed in each $\mathcal{H}_m$ and $\Omega$ is the (direct) sum of $\Omega_m$.
It should be noted that there is no need for the index set of the dictionary in each RKHS to equate each other~\cite{ishida2013}.
Therefore, the output of our filter in \eqref{eq:proposed_model} can be rewritten as a multikernel adaptive filter with $r^{(n)}$ different kernels:
\begin{align}
\nonumber
y^{(n)}&=\langle \Omega^{(n)}, \kappa(\cdot,\bm{u}^{(n)})\rangle_{\mathcal{H}}\\
       &=\underset{j\in\mathcal{J}^{(n)}}{\sum}\langle \Omega_j^{(n)}, \kappa(\cdot, \bm{u}^{(n)}; \bm{Z}_j^{(n)})\rangle_{\mathcal{H}_j},
\end{align}
where $\Omega_j^{(n)} = h_j^{(n)}\kappa(\cdot, \bm{c}_j^{(n)}; \bm{Z}_j^{(n)})$.
\subsection{Center Vectors Update}\label{update_matrix_center}
The update rule for each center vector can be derived by using a LMS algorithm:
\begin{align}
\bm{c}_{j}^{(n+1)} = \bm{c}_{j}^{(n)} -\eta_{\rm c}\left.\frac{\partial J^{(n)}(\bm{c}_j)}{\partial \bm{c}_j}\right|_{\bm{c}_j=\bm{c}_{j}^{(n)}}\label{eq:center}
\end{align}
%-
where $\eta_{\rm c}>0$ denotes a step size and
\begin{align}
\left.\frac{\partial J^{(n)}(\bm{c}_j)}{\partial \bm{c}_j}\right|_{\bm{c}_j=\bm{c}_{j}^{(n)}}&=-2 e^{(n)} h_{j}^{(n)}\kappa(\bm{u}^{(n)},\bm{c}^{(n)}_{j};\bm{Z}^{(n)}_j)\times\nonumber\\
& (\bm{Z}_j^{(n)}+{\bm{Z}_j^{(n)}}^{\top})(\bm{u}^{(n)}-\bm{c}_{j}^{(n)})
\end{align}
It should be remarked that the update rules of center vectors for the standard Gaussian kernel adaptive filters by using LMS algorithm have been also proposed in \cite{saide2012,saide2013,ishida2015}.
\subsection{Precision Matrices Update}\label{update_matrix_width}
\begin{figure}[t]
    \vspace{18pt}
	\subfloat[Matrix exponentiated gradient (MEG) update. This rule requires the computation of $\log \bm{Z}$ that can be unstable when the eigenvalues of $\bm{Z}$ are very small.]{\includegraphics[width=\linewidth]{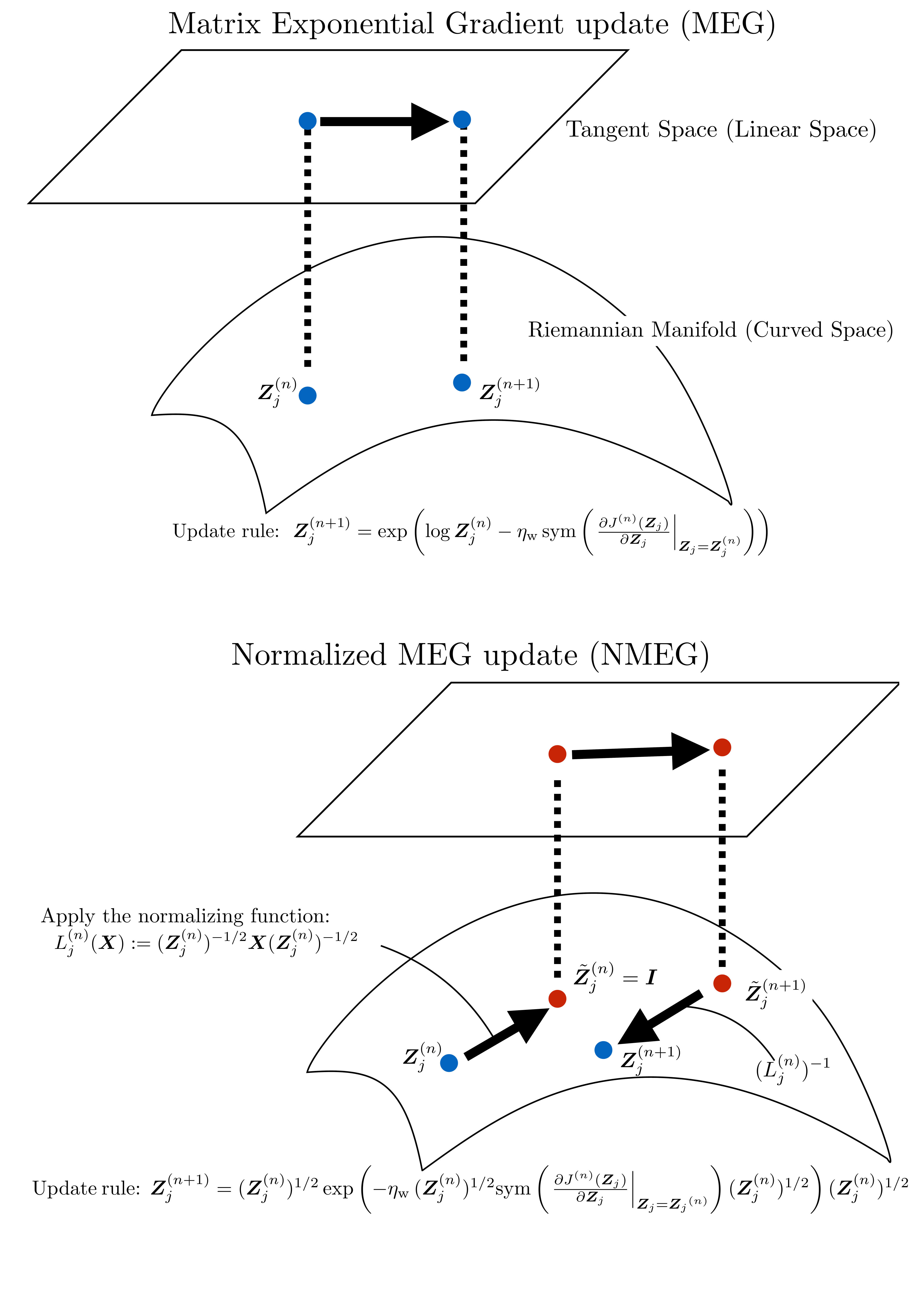}\label{fig:mse05}}\\
	
	\vspace{18pt}
	\subfloat[Normalized MEG (NMEG) update. This rule can avoid the problem of MEG by using  normalization.]{\includegraphics[width=\linewidth]{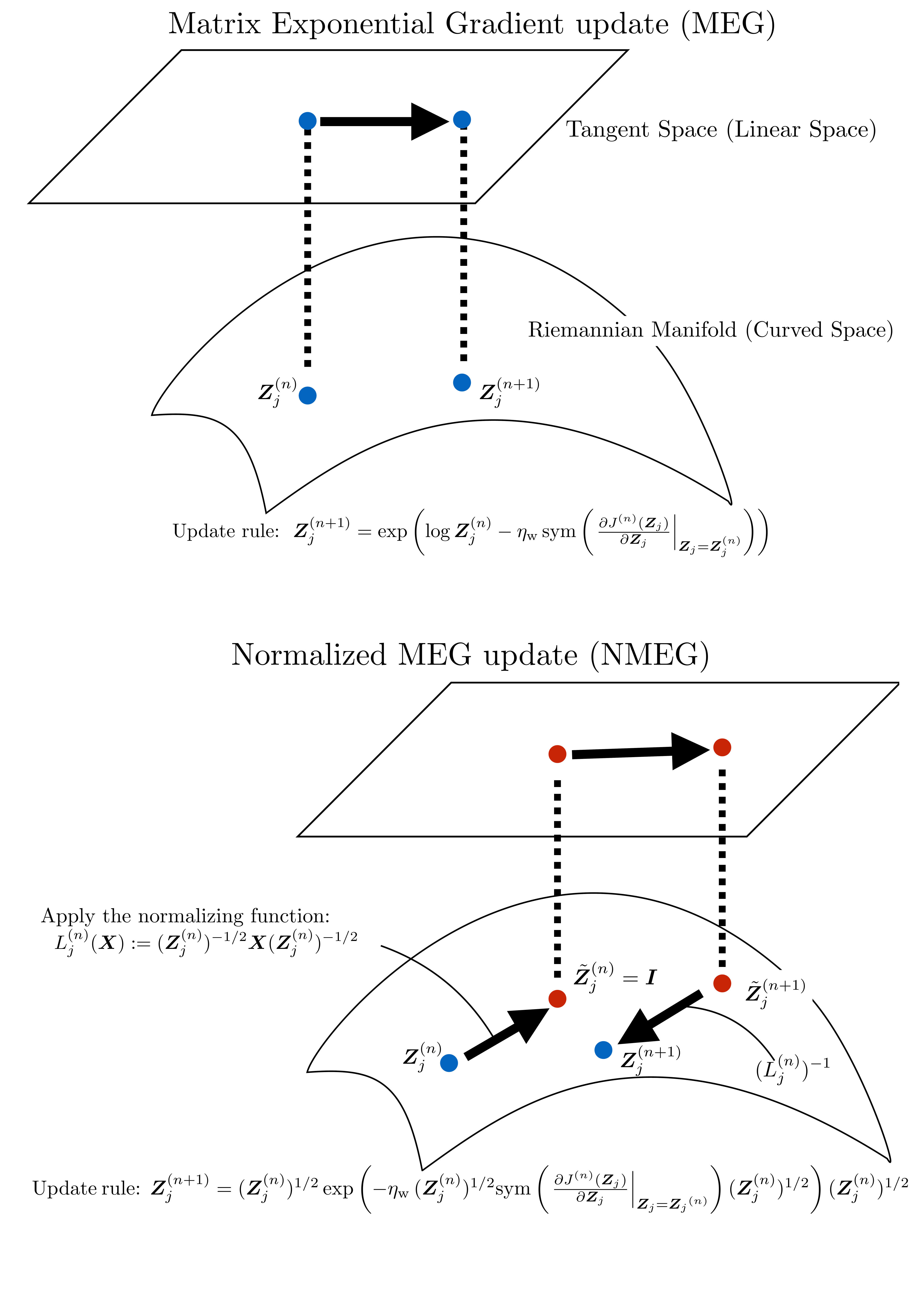}\label{fig:mse2}}
	\caption{Conceptual diagrams of (a) MEG and (b) NMEG.}
	\label{fig:MEG_NMEG}
	%\vspace{-0.8cm}
\end{figure}

In order to update the precision matrices, we consider two types of data-driven adaptation methods.
One is to apply the update rule for SPD matrices~\cite{tsuda2005} to updating the precision matrices in the kernel adaptive filtering.
The other is a novel update rule  for the precision matrices, where an effective normalization is employed.
This is the main contribution in the proposed method.
The Figure~\ref{fig:MEG_NMEG} illustrates these update rules.
\subsubsection{Matrix Exponentiated Gradient Update (MEG)}
To update precision matrices in the dictionary $\{\bm{Z}_j\}_{j \in\mathcal{J}^{(n)}}$ while preserving the SPD structure, the matrix exponentiated gradient (MEG) update~\cite{tsuda2005} is applied.
The update rule for $\bm{Z}_j$ can be derived to minimize the loss function in \eqref{eq:loss_matrix}:
\begin{align}
\nonumber
&{\bm{Z}}_{j}^{(n+1)}=\\
&\exp\left(\log{\bm{Z}}_{j}^{(n)}-\eta_{\rm w}\,{\rm sym}\left(\left.\frac{\partial J^{(n)}({\bm{Z}}_j)}{\partial {\bm{Z}}_j}\right|_{{\bm{Z}}_j={\bm{Z}}_{j}^{(n)}}\right)\right),
\label{eq:MEG}
\end{align}
where $\eta_{\rm w}>0$ denotes a step size and
\begin{align}
&\left.\frac{\partial J^{(n)}({\bm{Z}}_j)}{\partial {\bm{Z}}_j}\right|_{{\bm{Z}}_j={\bm{Z}}_{j}^{(n)}}\nonumber\\
&=2e^{(n)}{h_{j}^{(n)}}\kappa(\bm{u}^{(n)},\bm{c}^{(n)}_{j};\bm{Z}^{(n)}_j)(\bm{u}^{(n)} - \bm{c}_j)(\bm{u}^{(n)} - \bm{c}_j)^{\top}.
\end{align}
For a square matrix $\bm{X}$, ${\rm sym}(\bm{X}):=(\bm{X}+\bm{X}^{\top})/2$ denotes the symmetric part of $\bm{X}$, while $\exp(\bm{X})$ and $\log(\bm{X})$ denote matrix exponential and principal matrix logarithm, respectively~\cite{tsuda2005}.
\subsubsection{Normalized Matrix Exponentiated Gradient Update (NMEG)}
Even though the MEG can update each precision matrix $\bm{Z}$ while preserving its SPD structure,
the computation of $\log \bm{Z}$ can be unstable when the eigenvalues of $\bm{Z}$ are too close to zero\footnote{A symmetric positive-definite matrix $\bm{Z}$ with $L$ all-distinct eigenvalues may be decomposed as $\bm{W}\mathrm{diag}(\lambda_1,\lambda_2,\ldots,\lambda_L)\bm{W}^\top$, with $\bm{W}$ orthogonal. Therefore, $\log\bm{Z}=\bm{W}\mathrm{diag}(\log\lambda_1,\log\lambda_2,\ldots,\log\lambda_L)\bm{W}^\top$: If an eigenvalue gets too close to zero, the matrix logarithm becomes numerically unstable. In general, a matrix logarithm is well-defined only in a neighbor of the identity matrix $\bm{I}$.}.
To overcome this problem, the following normalizing function by the current value $\bm{Z}_{j}^{(n)}$ is proposed:
\begin{align}
L_{j}^{(n)}(\bm{X}):=(\bm{Z}_{j}^{(n)})^{-1/2}\bm{X}(\bm{Z}_{j}^{(n)})^{-1/2}.\label{eq:left_function}
\end{align}
Since each precision matrix $\bm{Z}_{j}^{(n)}$ is symmetric and positive-definite, their always inverse exists and their matrix square root returns a symmetric, real-valued matrix. The inverse (de-normalizing) function is
\begin{align}
(L_{j}^{(n)})^{-1}(\bm{X}):=(\bm{Z}_{j}^{(n)})^{1/2}\bm{X}(\bm{Z}_{j}^{(n)})^{1/2}.\label{eq:right_function}
\end{align}
We define the normalized precision matrix $\bm{Z}$ by \eqref{eq:left_function} as
$\tilde{\bm{Z}}:=L_{j}^{(n)}(\bm{Z})$.
If we apply the MEG update to $\tilde{\bm{Z}}_{j}^{(n)}$ instead of $\bm{Z}_{j}^{(n)}$,
we get the update rule
\begin{align}
\nonumber
&\tilde{\bm{Z}}_{j}^{(n+1)}\\
&=\exp\left(\log \tilde{\bm{Z}}_{j}^{(n)}-\eta_{\rm w}\,{\rm sym}\left(\left.\frac{\partial {J^{(n)}}({\bm{Z}}_j)}{\partial \tilde{\bm{Z}}_{j}}\right|_{\tilde{\bm{Z}}_j={\tilde{\bm{Z}}_{j}}^{(n)}}\right)\right),\label{eq:updateLambda}
\end{align}
where $\bm{Z}_j={\bm{Z}}_j(\tilde{\bm{Z}}_{j})$ is to be thought of as a compound function, in fact, it holds that ${\bm{Z}}_j(\tilde{\bm{Z}}_{j}):=(L_{j}^{(n)})^{-1}(\tilde{\bm{Z}}_{j})$.
Notice that $\tilde{\bm{Z}}_{j}^{(n)}$ can be written as
\begin{align}
\tilde{\bm{Z}}_{j}^{(n)} = L_{j}^{(n)}({\bm{Z}_j^{(n)}})=(\bm{Z}_{j}^{(n)})^{-1/2}\bm{Z}_{j}^{(n)}(\bm{Z}_{j}^{(n)})^{-1/2}=\bm{I},
\end{align}
where $\bm{I}\in\mathbb{R}^{L\times L}$ is an identity matrix.
Since $\log\bm{I}=\bm{0}$, the update rule \eqref{eq:updateLambda} can be written as
\begin{align}
\nonumber
&\tilde{\bm{Z}}_{j}^{(n+1)}\\
&=\exp\left(-\eta_{\rm w}{\rm sym}\left(\left.\frac{\partial {J^{(n)}}({\bm{Z}}_j(\tilde{\bm{Z}}_{j}))}{\partial \tilde{\bm{Z}}_{j}}\right|_{\tilde{\bm{Z}}_j={\tilde{\bm{Z}}_{j}}^{(n)}}\right)\right),\label{eq:updateLambda_2}
\end{align}
To find the derivative of function $J^{(n)}({\bm{Z}_{j}})$ with respect to $\tilde{\bm{Z}}_{j}$,
the following chain rule~\cite{petersen2012matrix} is used:
\begin{align}
\nonumber
&\left(\frac{\partial {J^{(n)}}({\bm{Z}}_j(\tilde{\bm{Z}}_{j}))}{\partial \tilde{\bm{Z}}_{j}}\right)_{kl}\\
&={\rm Tr}\left[\left(\frac{\partial {J^{(n)}}({\bm{Z}}_j)}{\partial {\bm{Z}}_j}\right)^{\top}\frac{\partial {\bm{Z}}_j}{\partial (\tilde{\bm{Z}}_{j})_{kl}}\right]\nonumber\\
&={\rm Tr}\left[\left(\frac{\partial {J^{(n)}}({\bm{Z}}_j)}{\partial {\bm{Z}}_j}\right)^{\top}(\bm{Z}_{j}^{(n)})^{1/2}\frac{\partial \tilde{\bm{Z}}_{j}}{\partial (\tilde{\bm{Z}}_{j})_{kl}}(\bm{Z}_{j}^{(n)})^{1/2}\right]\nonumber\\
&={\rm Tr}\left[\left(\frac{\partial {J^{(n)}}({\bm{Z}}_j)}{\partial {\bm{Z}}_j}\right)^{\top}(\bm{Z}_{j}^{(n)})^{1/2}\bm{S}_{kl}(\bm{Z}_{j}^{(n)})^{1/2}\right]\nonumber\\
&=\left((\bm{Z}_{j}^{(n)})^{1/2}\left(\frac{\partial {J^{(n)}}({\bm{Z}}_j)}{\partial {\bm{Z}}_j}\right)^{\top}(\bm{Z}_{j}^{(n)})^{1/2}\right)_{lk},\label{eq:chain_rule}
\end{align}
where the notation $(\bm{X})_{kl}$ denotes again the $(k,l)$-th entry of a matrix $\bm{X}$, $\rm{Tr}(\cdot)$ denotes matrix trace, and $\bm{S}_{kl}$ is the single-entry matrix~\cite{petersen2012matrix}, whose $(k,l)$-th entry is $1$ and each other entry takes the $0$ value.
From the property \eqref{eq:chain_rule}, we get
\begin{align}
\frac{\partial {J^{(n)}}({\bm{Z}}_j(\tilde{\bm{Z}}_{j}))}{\partial \tilde{\bm{Z}}_{j}}
&=\left((\bm{Z}_{j}^{(n)})^{1/2}\left(\frac{\partial {J^{(n)}}({\bm{Z}}_j)}{\partial {\bm{Z}}_j}\right)^{\top}(\bm{Z}_{j}^{(n)})^{1/2}\right)^{\top}\nonumber\\
&=(\bm{Z}_{j}^{(n)})^{1/2}\frac{\partial {J^{(n)}}({\bm{Z}}_j)}{\partial {\bm{Z}}_j}(\bm{Z}_{j}^{(n)})^{1/2},\label{eq:derivative}
\end{align}
thanks to the symmetry of the involved matrices and expressions.
Using the formula \eqref{eq:derivative}, the update rule \eqref{eq:updateLambda_2} can be written as
\begin{align}
\nonumber
&\tilde{\bm{Z}}_{j}^{(n+1)}\\
=&\exp\left(-\eta_{\rm w}\,{\rm sym}\left((L_{j}^{(n)})^{-1}\left(\left.\frac{\partial {J^{(n)}}({\bm{Z}}_j)}{\partial {\bm{Z}}_j}\right|_{{\bm{Z}}_j={{\bm{Z}}_{j}}^{(n)}}\right)\right)\right)\nonumber\\
=&\exp\left(-\eta_{\rm w}\,(L_{j}^{(n)})^{-1}\left({\rm sym}\left(\left.\frac{\partial {J^{(n)}}({\bm{Z}}_j)}{\partial {\bm{Z}}_j}\right|_{{\bm{Z}}_j={{\bm{Z}}_{j}}^{(n)}}\right)\right)\right).\label{eq:updateLambda_new}
\end{align}
Thanks to the normalizing function, we can update the precision matrices stably on the tangent space at identity.
Then, the $(n+1)$-th precision matrix is obtained by applying the inverse function.
Therefore, the update rule \eqref{eq:NMEG} is derived.
\begin{figure*}
\begin{align}
&\bm{Z}_{j}^{(n+1)}
=(L_{j}^{(n)})^{-1}(\tilde{\bm{Z}}_{j}^{(n+1)})
=(\bm{Z}_{j}^{(n)})^{1/2}\tilde{\bm{Z}}_{j}^{(n+1)}(\bm{Z}_{j}^{(n)})^{1/2}\nonumber\\
&=(\bm{Z}_{j}^{(n)})^{1/2}\exp\left(-\eta_{\rm w}\,(\bm{Z}_{j}^{(n)})^{1/2}{\rm sym}\left(\left.\frac{\partial {J^{(n)}}({\bm{Z}}_j)}{\partial {\bm{Z}}_j}\right|_{{\bm{Z}}_j={{\bm{Z}}_{j}}^{(n)}}\right)(\bm{Z}_{j}^{(n)})^{1/2}\right)(\bm{Z}_{j}^{(n)})^{1/2}.\label{eq:NMEG}
\end{align}
\end{figure*}
From \eqref{eq:NMEG}, we can see that unlike \eqref{eq:MEG}, this update rule dose not require the computation of $\log \bm{Z}$.
We call this update rule the normalized matrix exponentiated gradient (NMEG) update.

As a special instance, let us consider the case $L=1$.
%Above NMEG update rule runs when the size of $\bm{Z}$ is not unity.
The NMEG update rule in the case of $L=1$ can be derived by replacing a precision matrix $\bm{Z}$ with a scalar parameter $\zeta>0$ in \eqref{eq:NMEG}:
\begin{align}
&\zeta_{j}^{(n+1)}
%\nonumber\\
%&=\sqrt{\zeta_{j}^{(n)}}\exp\left(-\eta_{\rm w}\,\sqrt{\zeta_{j}^{(n)}}\left(\left.\frac{\partial
%{J^{(n)}}({\zeta}_j)}{\partial {\zeta}_j}\right|_{{\zeta}_j={{\zeta}_{j}}^{(n)}}\right)
%\sqrt{\zeta_{j}^{(n)}}\right)\sqrt{\zeta_{j}^{(n)}}\nonumber\\
&=\zeta_j^{(n)}\exp\left(-\eta_{\rm w}\zeta^{(n)}_j\left.\frac{\partial J^{(n)}(\zeta_j)}{\partial \zeta_j}\right|_{\zeta_j=\zeta^{(n)}_j}\right),\label{eq:NEG}
\end{align}
which apparently keeps each parameter $\zeta_j$ in the positive half-line.
The partial derivative of the cost function reads
\begin{align}\label{wada_rule}
&\left.\frac{\partial J^{(n)}(\zeta_j)}{\partial \zeta_j}\right|_{\zeta_j=\zeta^{(n)}_j}\nonumber\\
&=-2\zeta^{(n)}_j e^{(n)}h_{j}^{(n)}\kappa(\bm{u}^{(n)},\bm{c}^{(n)}_j; \zeta^{(n)}_j)\|\bm{u}^{(n)}-\bm{c}_j^{(n)}\|^2.
\end{align}
Such special case was proposed and discussed in the recent contributions~\cite{wada2017,wada2018}.

The update rule~\eqref{eq:NMEG} was derived on the basis of matrix normalization, therefore, it is legitimate to wonder if it constitutes a valid algorithm to update a matrix in the space of SPD tensors. The answer is positive, indeed, since the rule~\eqref{eq:NMEG} may be regarded as an application of a general geodesic-based stepping rule on the Lie group of symmetric positive-definite matrices induced by the canonical metrics, namely
\begin{align}
&\bm{Z}_{j}^{(n+1)}\nonumber\\
&=g_{\bm{Z}_{j}^{(n)}}\left(-\eta_{\rm w}\,\bm{Z}_{j}^{(n)}{\rm sym}\left(\left.\frac{\partial {J^{(n)}}({\bm{Z}}_j)}{\partial {\bm{Z}}_j}\right|_{{\bm{Z}}_j={{\bm{Z}}_{j}}^{(n)}}\right)\bm{Z}_{j}^{(n)}\right),\label{eq:GBS}
\end{align}
where the function $g_X(V)$ denotes a geodesic arc in the SPD space departing from a point $X$ in the direction $V$ and is given by
\begin{align}
g_{X}(V):=X^{1/2}\exp(X^{-1/2}VX^{-1/2})X^{1/2},\label{eq:Garc}
\end{align}
as explained, for example, in \cite[Eq. (22)]{fiori2009} and \cite[Eq. (3.6)]{ustf2017}. Notice, in addition, that the argument of the function $g$ in \eqref{eq:GBS} is proportional to the Riemannian gradient of the criterion function $J$ with respect to the canonical metrics.
\subsection{$\ell_1$-regularized KNLMS Incorporated With Generalized Gaussian Kernel Parameters}
To avoid overfitting and monotonic increase of the cardinality of a dictionary, the proposed update rules for the generalized Gaussian parameters are incorporated with the $\ell_1$-regularized least squares for updating the filter coefficients~\cite{gao2014}.
The proposed method is summarized in the Algorithm~\ref{alg1}.
\begin{algorithm}[t]
	\caption{Dictionary Learning for Generalized Gaussian Kernel Adaptive Filtering}
	\label{alg1}
	\begin{algorithmic}[1]
		\STATE Set the initial size of dictionary, $n=1$, and precision matrix of kernel, $\bm{Z}_{\rm init}$.
		\STATE Add $(\bm{u}^{(0)},\bm{Z}_{\rm init})$ into the dictionary as the $1$st member, $\mathcal{D}^{(0)}=\{\bm{u}^{(0)},\bm{Z}_{\rm init})\}$.
		\FOR{$n>1)$}
%		\WHILE{\hl{$\{\bm{u}^{(n)}, d^{(n)}\}\ (n>1)$ available [notation not clear]}}
		\STATE Add $(\bm{u}^{(n)},\bm{Z}_{\rm init})$ into the dictionary as the new member,
		$\mathcal{D}^{(n)}=\{\mathcal{D}^{(n-1)}, (\bm{u}^{(n)},\bm{Z}_{\rm init})\}$.
		\STATE \label{alg:Mcenter} Update the center vectors using \eqref{eq:center}
		\STATE \label{alg:Mwidth} Update the precision matrices using MEG in \eqref{eq:MEG} or NMEG in \eqref{eq:NMEG}.
		\STATE Update the filter coefficients using \eqref{eq:l1}.
		\FOR{$j$ such that $h_{j} = 0$}
		\STATE Remove the $j$-th element from the dictionary, $\mathcal{D}^{(n)}$.
		\ENDFOR
		\STATE $n \gets n+1$
		\ENDFOR
	\end{algorithmic}
\end{algorithm}
\section{Numerical Examples}\label{sec:example}
In this section, we compare the KNLMS-$\ell_1$~\cite{gao2014}, the NMEG ($L=1$)~\cite{wada2017,wada2018} in \eqref{eq:NEG},
the MEG in \eqref{eq:MEG}, and the NMEG in \eqref{eq:NMEG} through three types of simulations.
As described in Algorithm \ref{alg1}, the NMEG ($L=1$), the MEG, and the NMEG update center vectors and are incorporated with a $\ell_1$-regularized least squares for updating the filter coefficients.
The first simulation is a time series prediction in a toy model defined by Gaussian functions with scalar widths.
The second simulation is online prediction in a toy model defined by Gaussian functions with precision matrices.
The last simulation consists in online prediction of the state of a Lorenz chaotic system.
\subsection{Time Series Prediction in Toy Model Constructed by Standard Gaussian Functions}\label{ex:sg}
\begin{table}[t]
	\centering
	\caption{Parameters in experiment \ref{ex:sg}}
	%\vspace{-2pt}
	\begin{tabular}{c|c} \hline
		KNLMS-$\ell_1$&$\mu=0.09,~\rho=0.03,~\zeta=1$\\
		&$\lambda=1.0\times10^{-3},~\beta=0.1$\\\hline
		NMEG ($L=1$)&$\mu=0.09,~\rho=0.03,~\zeta_{\rm init.}=1.0,~\lambda=1.0\times10^{-3}$\\
		&$\beta=0.1,~\eta_{c}=1.0\times10^{-3},~\eta_{\rm w}=0.05$\\\hline
		MEG&$\mu=0.09,~\rho=0.03,~\bm{Z}_{\rm init}=\bm{I},\lambda=1.0\times10^{-3}$\\
		&$\beta=0.1,~\eta_{c}=1.0\times10^{-3},~\eta_{\rm w}=0.05,~L=2$\\\hline
		NMEG&$\mu=0.09,~\rho=0.03,~\bm{Z}_{\rm init}=\bm{I},\lambda=1.0\times10^{-3}$\\
		&$\beta=0.1,~\eta_{c}=1.0\times10^{-3},~\eta_{\rm w}=0.05,~L=2$\\\hline
	\end{tabular} \label{tb:ex:sg}
	%\vspace{-3pt}
\end{table}
\begin{figure}[t]
	\centering
	\includegraphics[width=\columnwidth,clip]{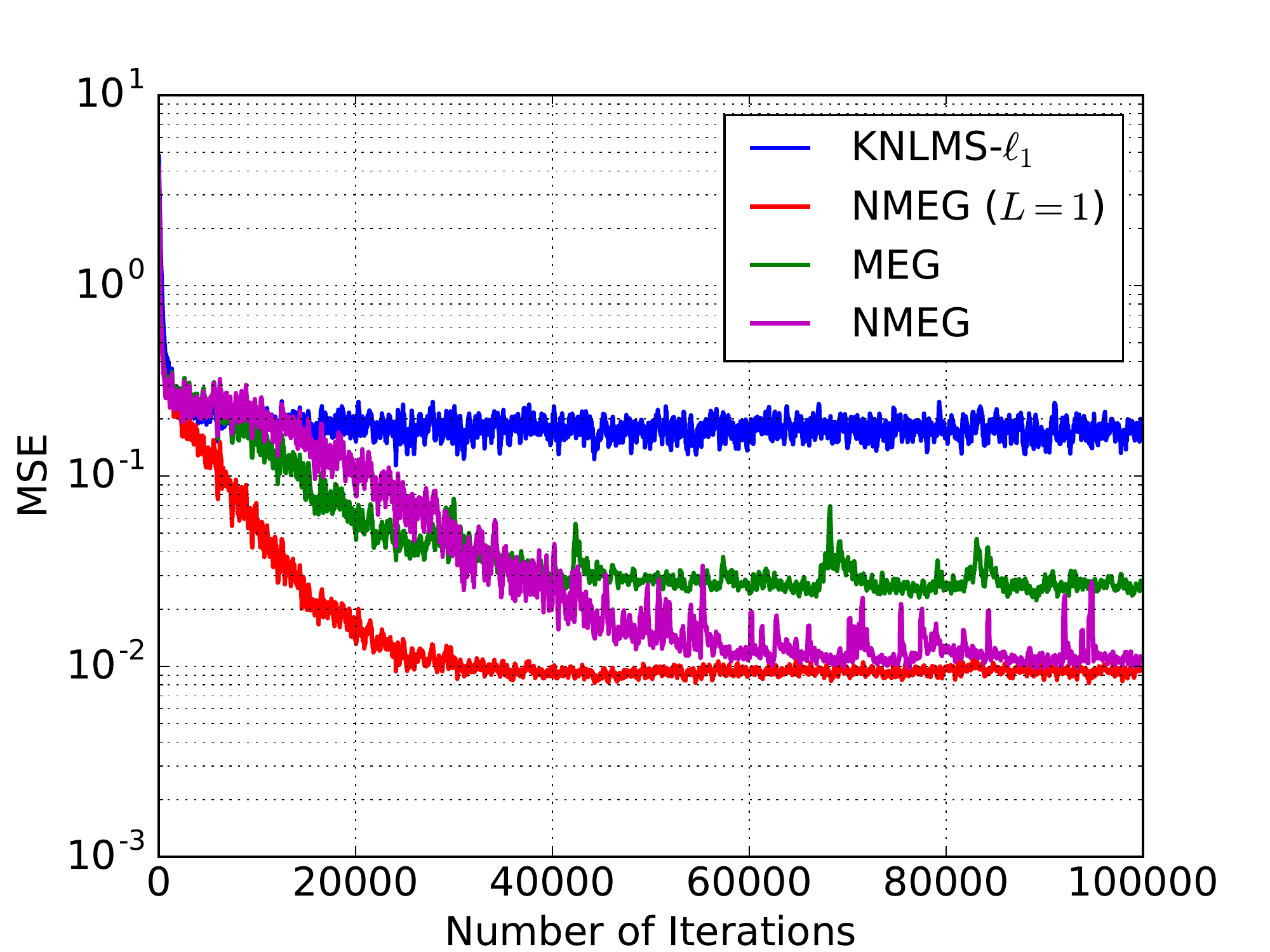}
	%\vspace{-2pt}
	\caption{Convergence curves of filters in experiment \ref{ex:sg}. These results were obtained as averages over 50 independent runs.}
	\label{fig:ex:sg_mse}
	%\vspace{-3pt}
\end{figure}
\begin{figure}[t]
	\centering
	\includegraphics[width=\columnwidth,clip]{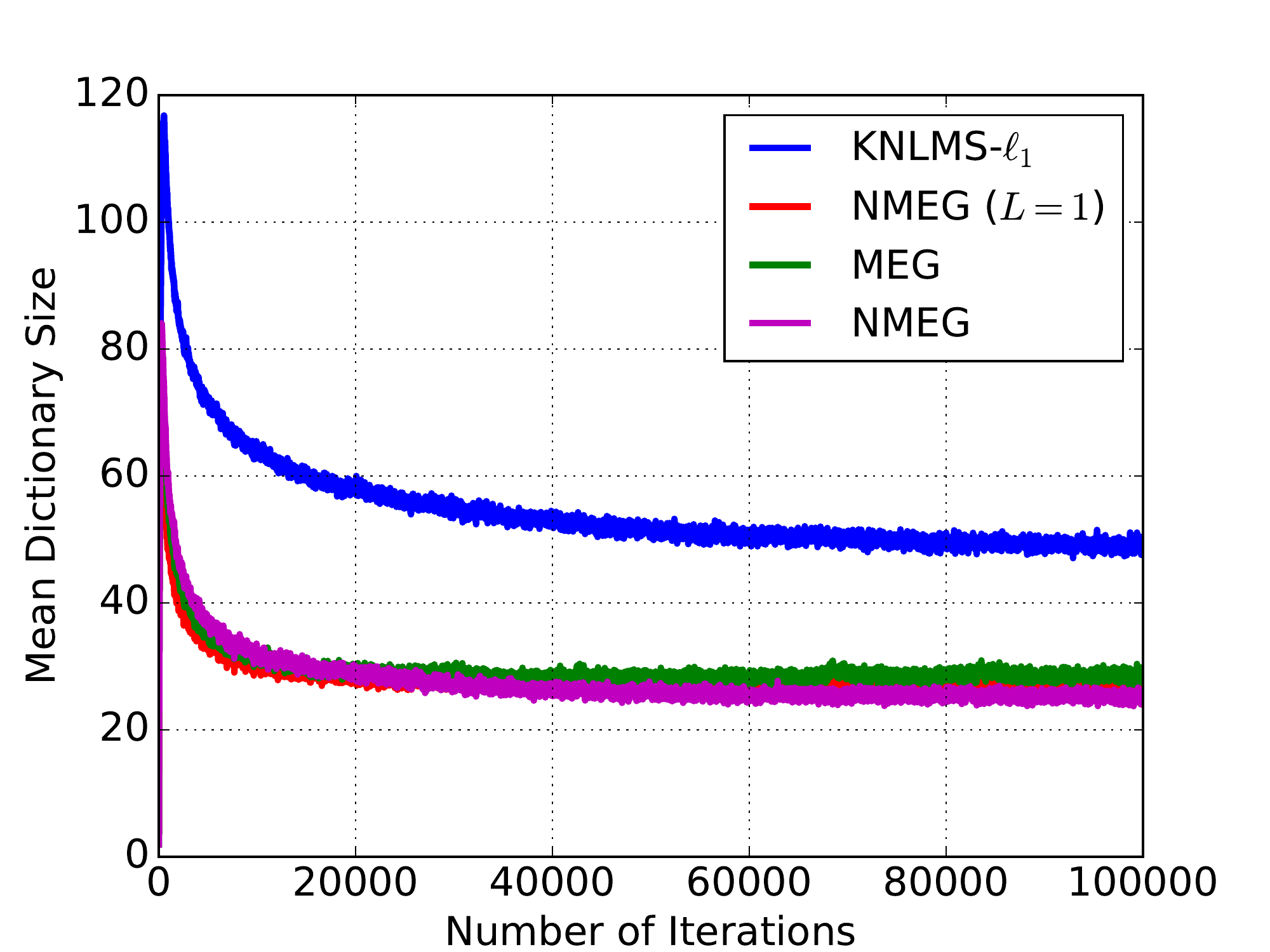}
	%\vspace{-2pt}
	\caption{Dictionary size evolution in experiment \ref{ex:sg}.These results were obtained as averages over 50 independent runs.}
	\label{fig:ex:sg_mds}
	%\vspace{-3pt}
\end{figure}
We consider the nonlinear system defined as follows:
\begin{align}
d^{(n)}&:=10\exp(-5\|\bm{u}^{(n)}-[3,3]^{\top}\|^2)+\nonumber\\
       &10\exp(-0.2\|\bm{u}^{(n)}-[7,7]^{\top}\|^2),
\end{align}
where $d^{(n)}$ is corrupted by noise sampled from a zero-mean Gaussian distribution with standard deviation equal to $0.3$.
The input signals $\bm{u}^{(n)}$ are sampled from a $2$-dimensional uniform distribution $[0,\ 10]\times[0,\ 10]$.
We adopted a mean squared error (MSE) measure as evaluation criterion.
The MSE is calculated by taking an arithmetic average over $50$ independent realizations.
The values of the parameters of the filters in this experiment are given in the Table \ref{tb:ex:sg}.
Figures~\ref{fig:ex:sg_mse} and~\ref{fig:ex:sg_mds} show the MSE and the mean dictionary size of filters at each iteration, respectively.
In the Figure \ref{fig:ex:sg_mse},
the NMEG ($L=1$), the MEG, and the NMEG show lower MSE than the KNLMS-$\ell_1$.
This implies the efficacy of updating width $\zeta$ or precision matrix $\bm{Z}$.
The NMEG ($L=1$) converges faster than other algorithms.
However, when $n$ is about $10,000$, the NMEG ($L=1$) and NMEG have almost the same MSE
even though the NMEG uses generalized Gaussian kernels.
The Figure~\ref{fig:ex:sg_mds} shows that
the NMEG ($L=1$), the MEG, and the NMEG keep a small dictionary size.
\subsection{Time Series Prediction in Toy Model Constructed by Generalized Gaussian Functions}\label{ex:gg}
\begin{table}[t]
	\centering
	\caption{Parameters in experiment \ref{ex:gg}}
	%\vspace{-2pt}
	\begin{tabular}{c|c} \hline
		KNLMS-$\ell_1$&$\mu=0.09,~\rho=0.03,~\zeta=1$\\
		&$\lambda=1.0\times10^{-3},~\beta=0.1$\\\hline
		NMEG ($L=1$)&$\mu=0.09,~\rho=0.03,~\zeta_{\rm init}=1.0,~\lambda=1.0\times10^{-3}$\\
		&$\beta=0.1,~\eta_{c}=1.0\times10^{-3},~\eta_{\rm w}=0.05$\\\hline
		MEG&$\mu=0.09,~\rho=0.03,~\bm{Z}_{\rm init}=\bm{I},\lambda=1.0\times10^{-3}$\\
		&$\beta=0.1,~\eta_{c}=1.0\times10^{-3},~\eta_{\rm w}=0.05,~L=2$\\\hline
		NMEG&$\mu=0.09,~\rho=0.03,~\bm{Z}_{\rm init}=\bm{I},\lambda=1.0\times10^{-3}$\\
		&$\beta=0.1,~\eta_{c}=1.0\times10^{-3},~\eta_{\rm w}=0.05,~L=2$\\\hline
	\end{tabular} \label{tb:ex:gg}
	%\vspace{-3pt}
\end{table}
\begin{figure*}[h!]
	\centering
	\subfloat[MSE when ${\bm{A}=\binom{\ 5 \ \ \ 0.5}{0.5 \ \ 0.2} }$]{\includegraphics[width=\columnwidth]{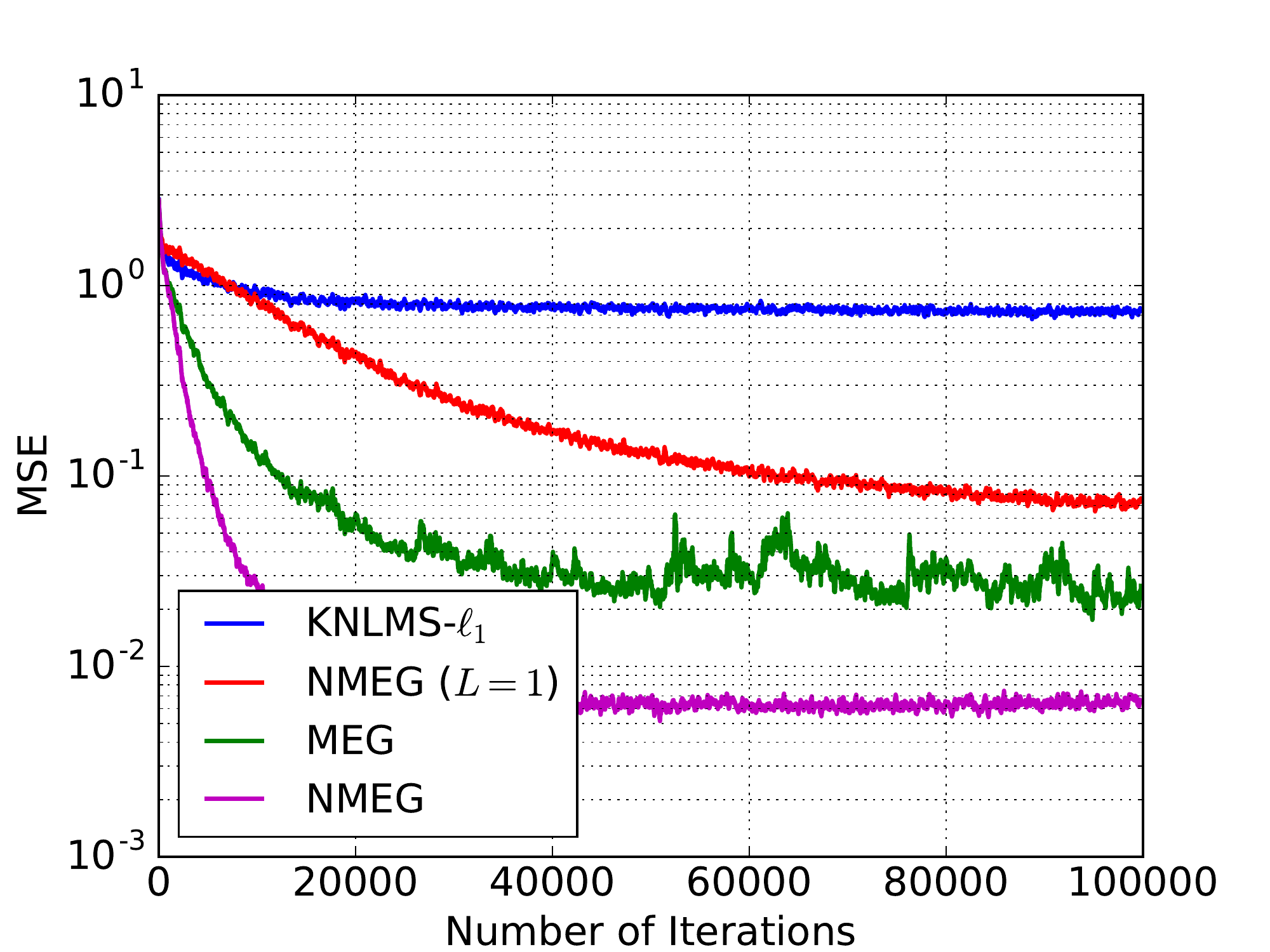}\label{fig:ex:gg_mse}}
	\subfloat[MSE when ${\bm{A}=\binom{\ 5 \ \ \ 0.5}{0.5 \ \ 10} }$]{\includegraphics[width=\columnwidth]{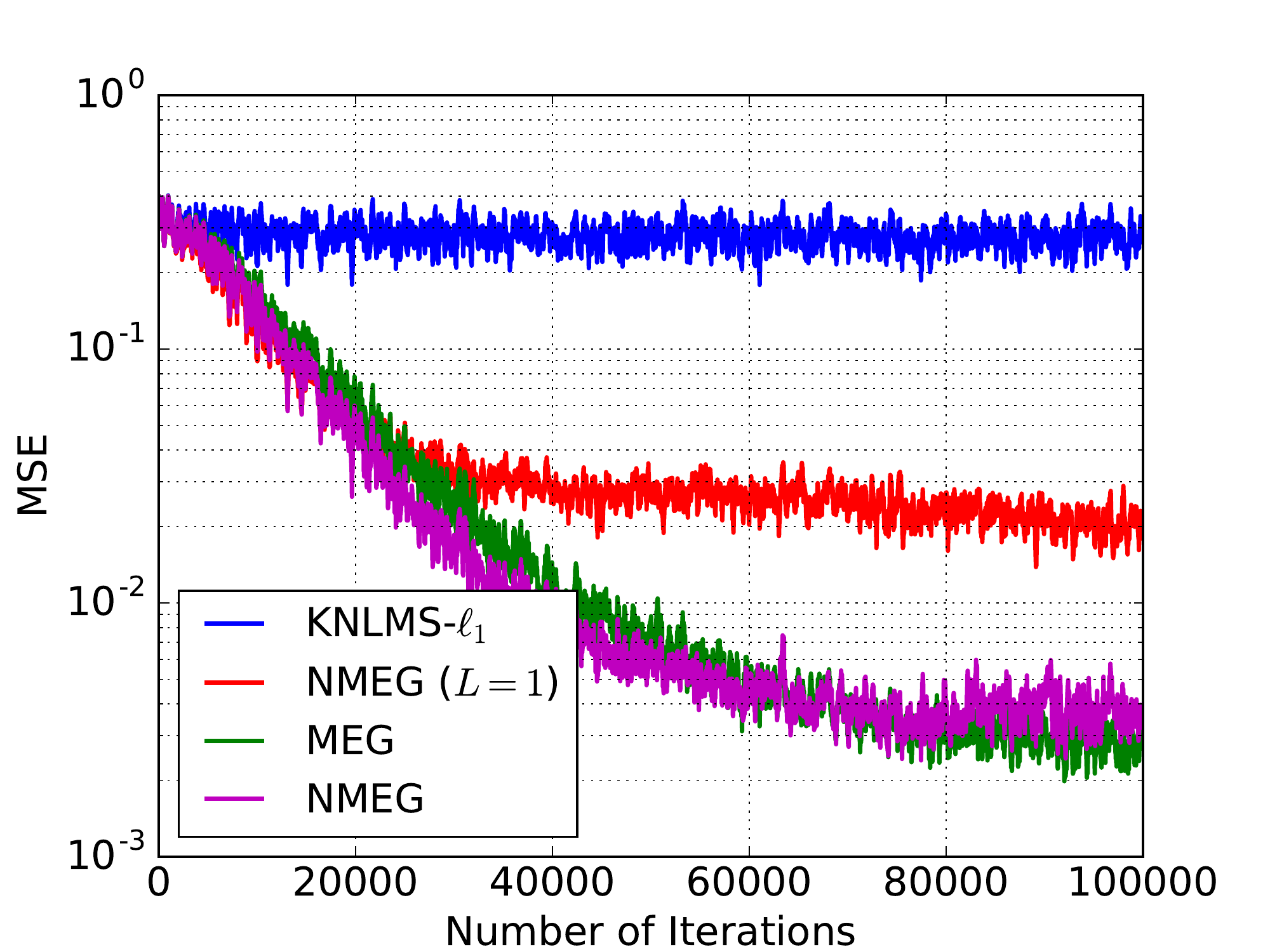}\label{fig:ex:gg2_mse}}\\
	%\vspace{-12pt}
	\subfloat[Mean dictionary size when ${\bm{A}=\binom{\ 5 \ \ \ 0.5}{0.5 \ \ 0.2} }$]{\includegraphics[width=\columnwidth]{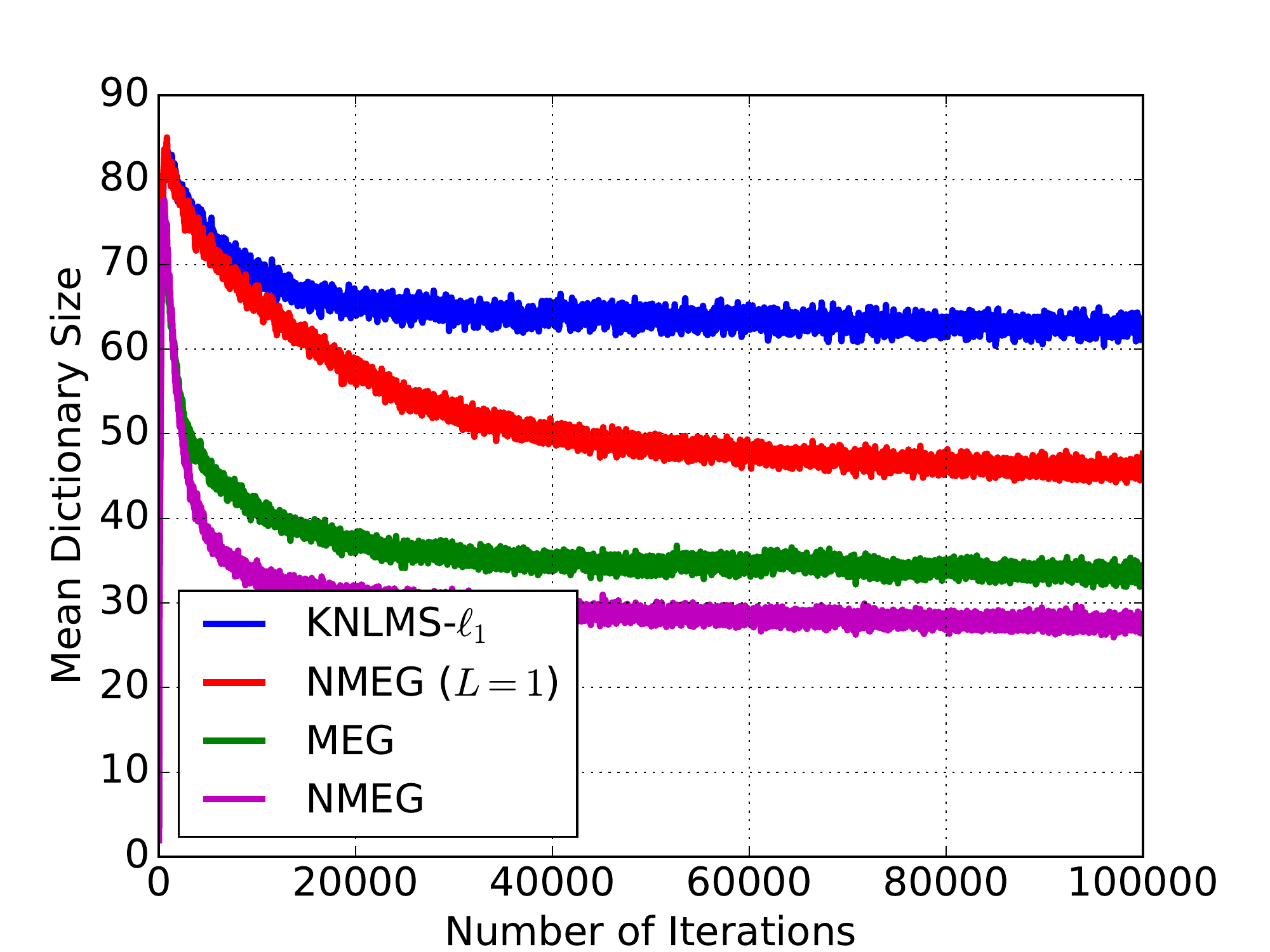}\label{fig:ex:gg_mds}}
	\subfloat[Mean dictionary size when ${\bm{A}=\binom{\ 5 \ \ \ 0.5}{0.5 \ \ 10}}$]{\includegraphics[width=\columnwidth]{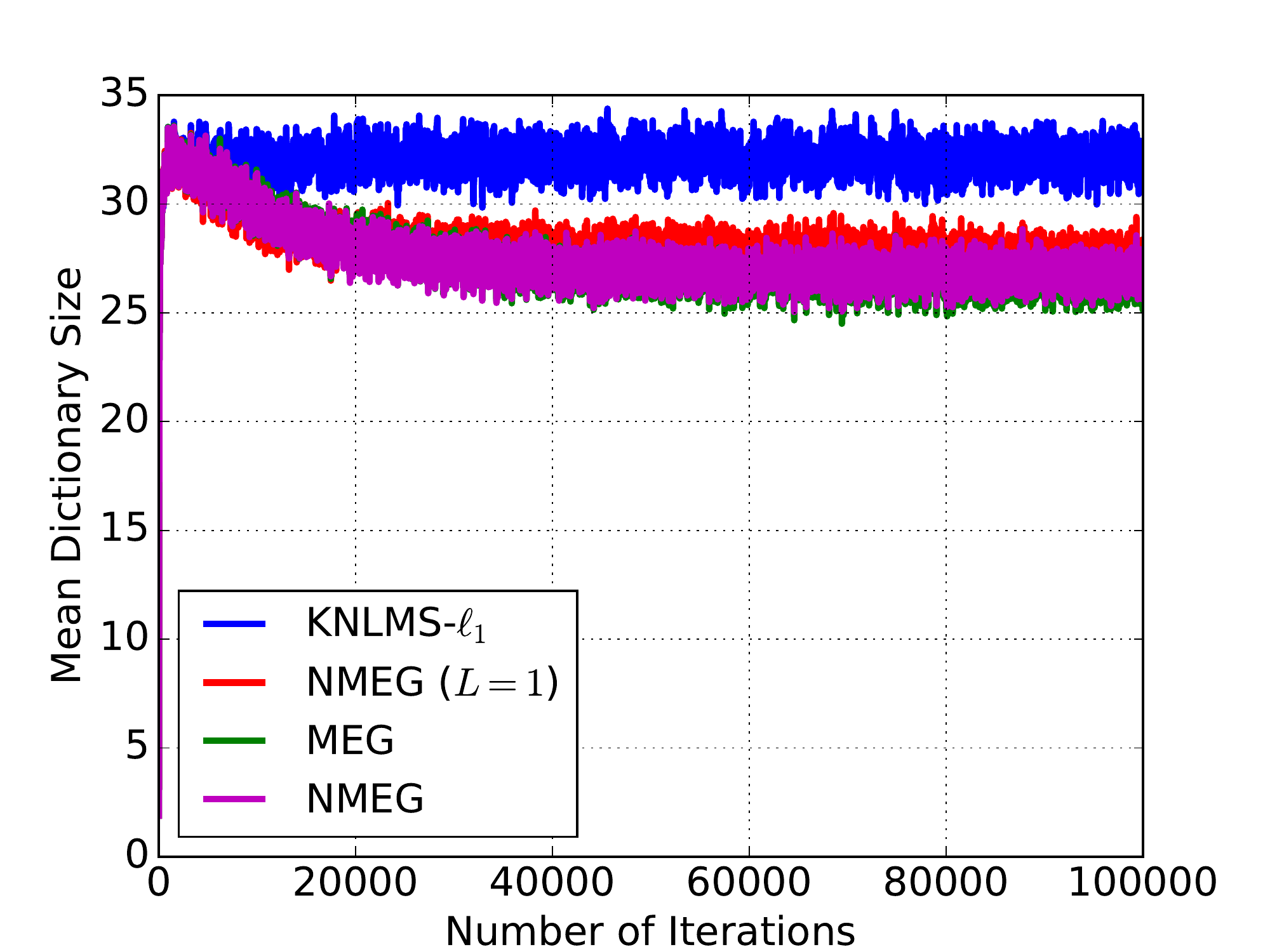}\label{fig:ex:gg2_mds}}\\

	\caption{Performance comparison: The learning curves of MSE ((a) and (b)) and mean dictionary size ((c) and (d)) for two different parameters $\bm{A}$.
		These results are calculated by taking an arithmetic average over $50$ independent realizations.}
	\label{fig:ex:gg}
	%\vspace{-4mm}
\end{figure*}
%
\if0
\begin{figure}[t]
	\centering
	\includegraphics[width=\columnwidth,clip]{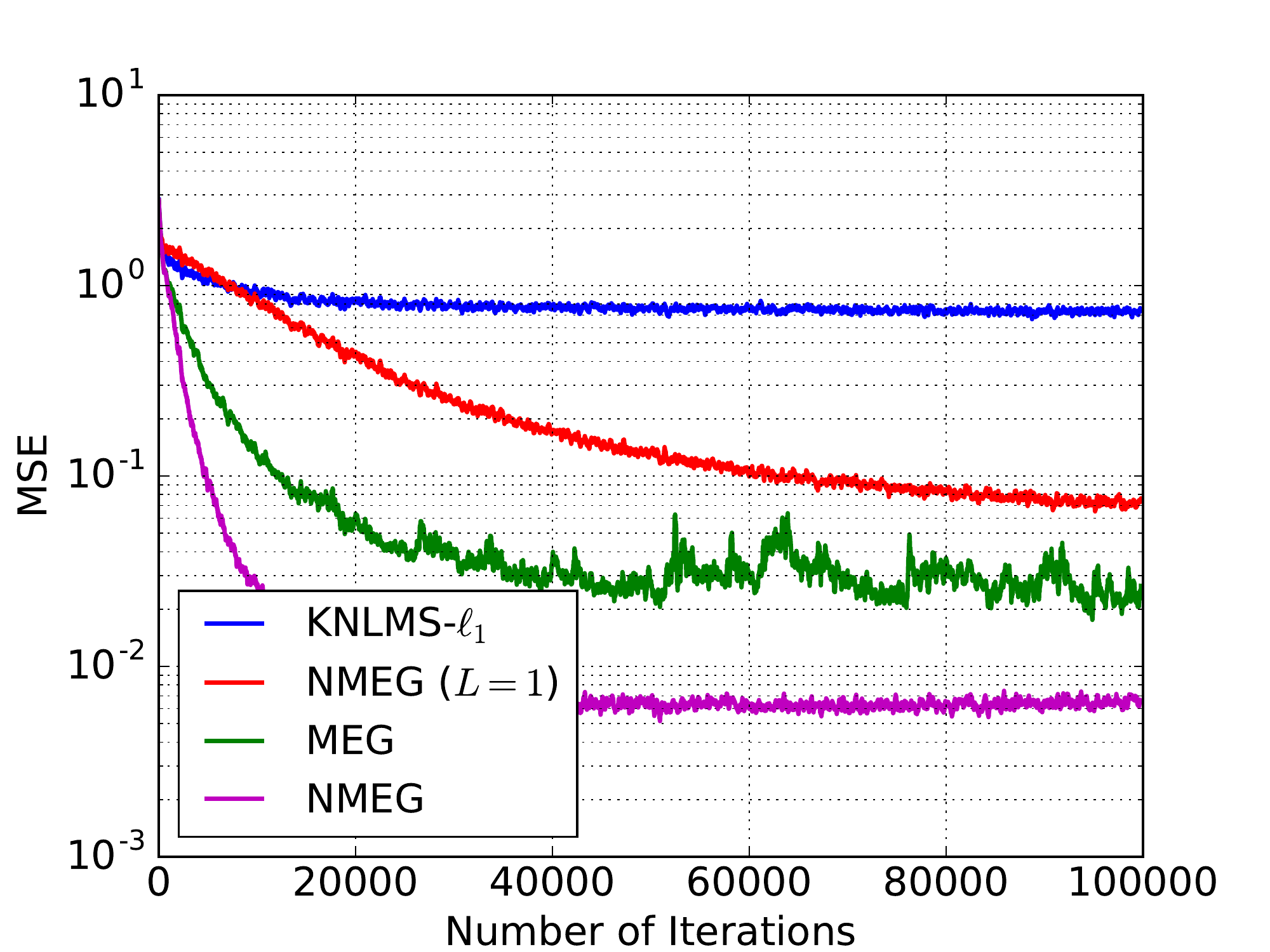}
	%\vspace{-2pt}
	\caption{Convergence curves of filters in experiment \ref{ex:gg}. These results were obtained as averages over 50 independent runs.}
	\label{fig:ex:gg_mse}
	%\vspace{-3pt}
\end{figure}
%
\begin{figure}[t]
	\centering
	\includegraphics[width=\columnwidth,clip]{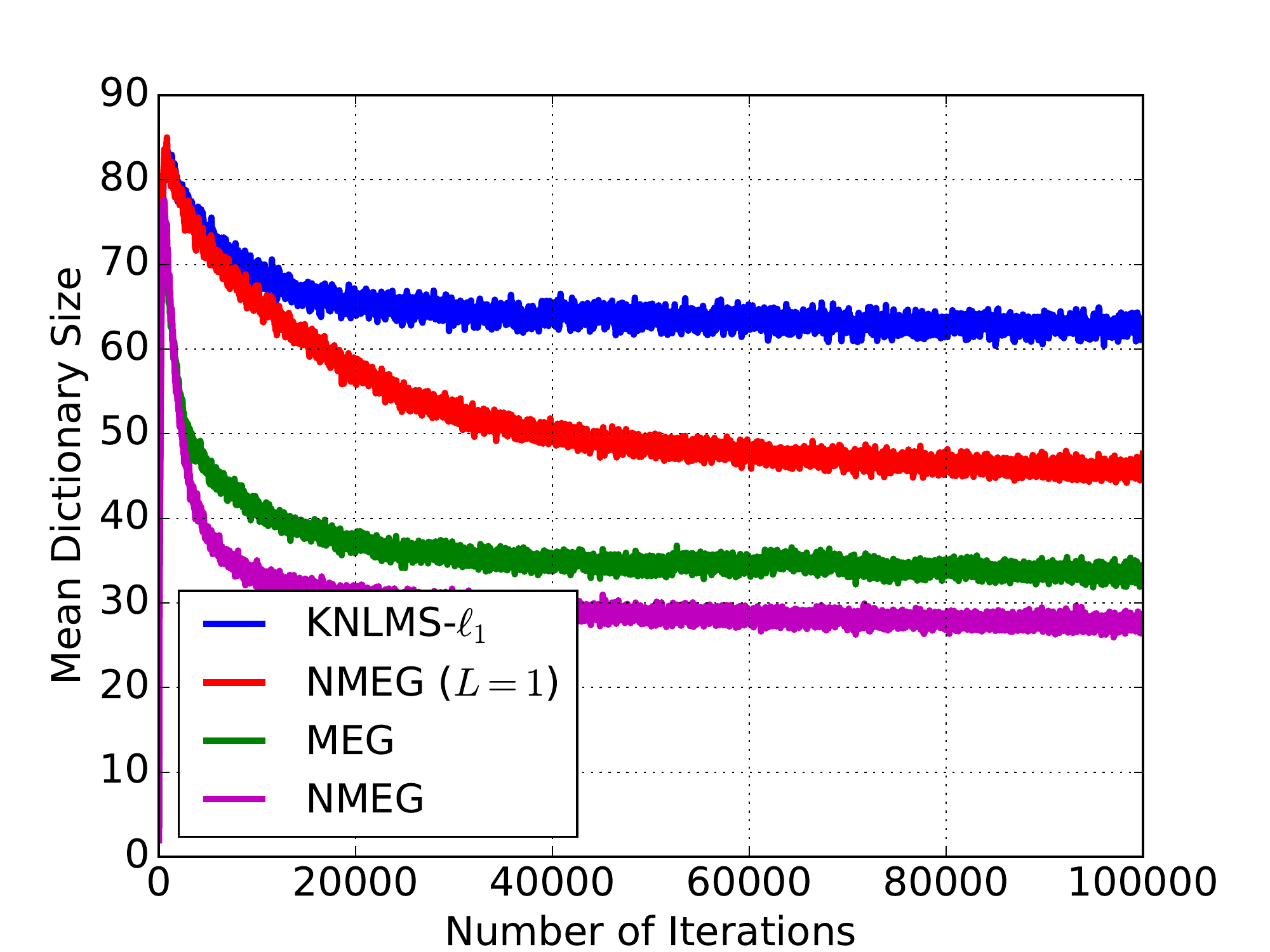}
	%\vspace{-2pt}
	\caption{Dictionary size evolution in experiment \ref{ex:gg}. These results were obtained as averages over 50 independent runs.}
	\label{fig:ex:gg_mds}
	%\vspace{-3pt}
\end{figure}
\fi
%
We consider the nonlinear system defined by:
\begin{align}
d^{(n)}&:=10\exp\left(-(\bm{u}^{(n)}-[3,3]^{\top})^{\top}\bm{A}
(\bm{u}^{(n)}-[3,3]^{\top})\right)\nonumber\\
&+10\exp\left(-(\bm{u}^{(n)}-[7,7]^{\top})^{\top}
\bm{A}(\bm{u}^{(n)}-[7,7]^{\top})\right),%\\
%\bm{A} &:= \begin{pmatrix} 5 & 0.5 \\ 0.5 & 0.2 \\ \end{pmatrix},
\end{align}
where $d^{(n)}$ is corrupted by a zero-mean Gaussian noise of standard deviation equal to $0.3$.
In the above system, $\bm{A}$ is a constant.
The input signals $\bm{u}^{(n)}$ are sampled from a $2$-dimensional uniform distribution $[0,\ 10]\times[0,\ 10]$.
The MSE is calculated by taking an arithmetic average over $50$ independent realizations.
Parameters in this experiment are given in the Table \ref{tb:ex:gg}.
We test two different cases of $\bm{A}$:
\begin{align}
 \begin{pmatrix} 5 & 0.5 \\ 0.5 & 0.2 \\ \end{pmatrix}, \begin{pmatrix} 5 & 0.5 \\ 0.5 & 10 \\ \end{pmatrix},
\end{align}
%
%The Figures~\ref{fig:ex:gg_mse} and~\ref{fig:ex:gg_mds} show the MSE and the mean dictionary size of filters at each iteration, respectively.
which have the smaller eivenvalues of $0.148$ and $4.95$, respectively.
The Figure~\ref{fig:ex:gg} shows the MSE and the mean dictionary size of filters at each iteration.
In Figures \ref{fig:ex:gg_mse} and \ref{fig:ex:gg2_mse}, the MEG and the NMEG show lower MSE than the KNLMS-$\ell_1$ and the NMEG ($L=1$).
This implies the efficacy of using (adaptive) generalized Gaussian kernels.
%Comparing the MEG with the NMEG, the NMEG shows lower MSE when $\bm{A}=\begin{pmatrix} 5 & 0.5 \\ 0.5 & 10 \\ \end{pmatrix}$ since the normalization can avoid a very small negative value caused by the logarithm map.
Comparing the MSE curves of the MEG and of the NMEG, it is immediate to see how the performance of the MEG algorithm degrades when the matrix $\bm{A}$ is close to zero, namely when $\bm{A}=\begin{pmatrix} 5 & 0.5 \\ 0.5 & 0.2 \\ \end{pmatrix}$, which implies that the term $\log \bm{Z}_j^{(n)}$ in \eqref{eq:MEG} is close to singularity, while the NMEG is able to perform well in both cases.
The Figures \ref{fig:ex:gg_mds} and \ref{fig:ex:gg2_mds} confirm that the NMEG requires the smallest dictionary size.
The above results support the efficacy of the proposed normalization for updating the precision matrix.
\subsection{Short-Term Chaotic Time-Series Prediction: Lorenz Chaotic System}\label{sec:lorenz}
\begin{figure}[t]
	\centering
	\includegraphics[width=\columnwidth,clip]{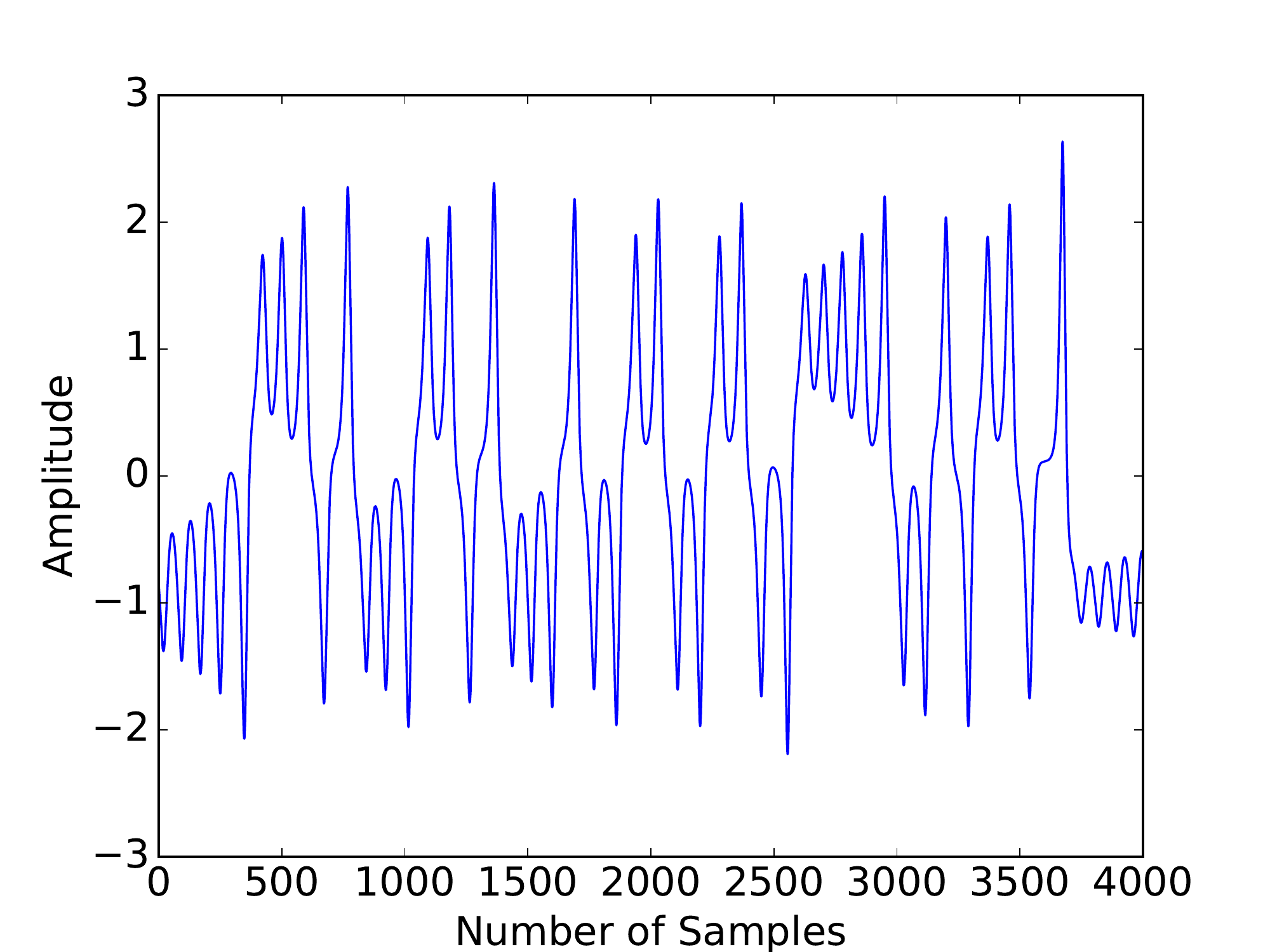}
	%\vspace{-2pt}
	\caption {Segment of the processed Lorenz time series.}
	\label{fig:lorenz}
	%\vspace{-3pt}
\end{figure}
\begin{figure*}
	\centering
	\subfloat[Tracking of KNLMS-$\ell_1$]{\includegraphics[width=0.5\linewidth]{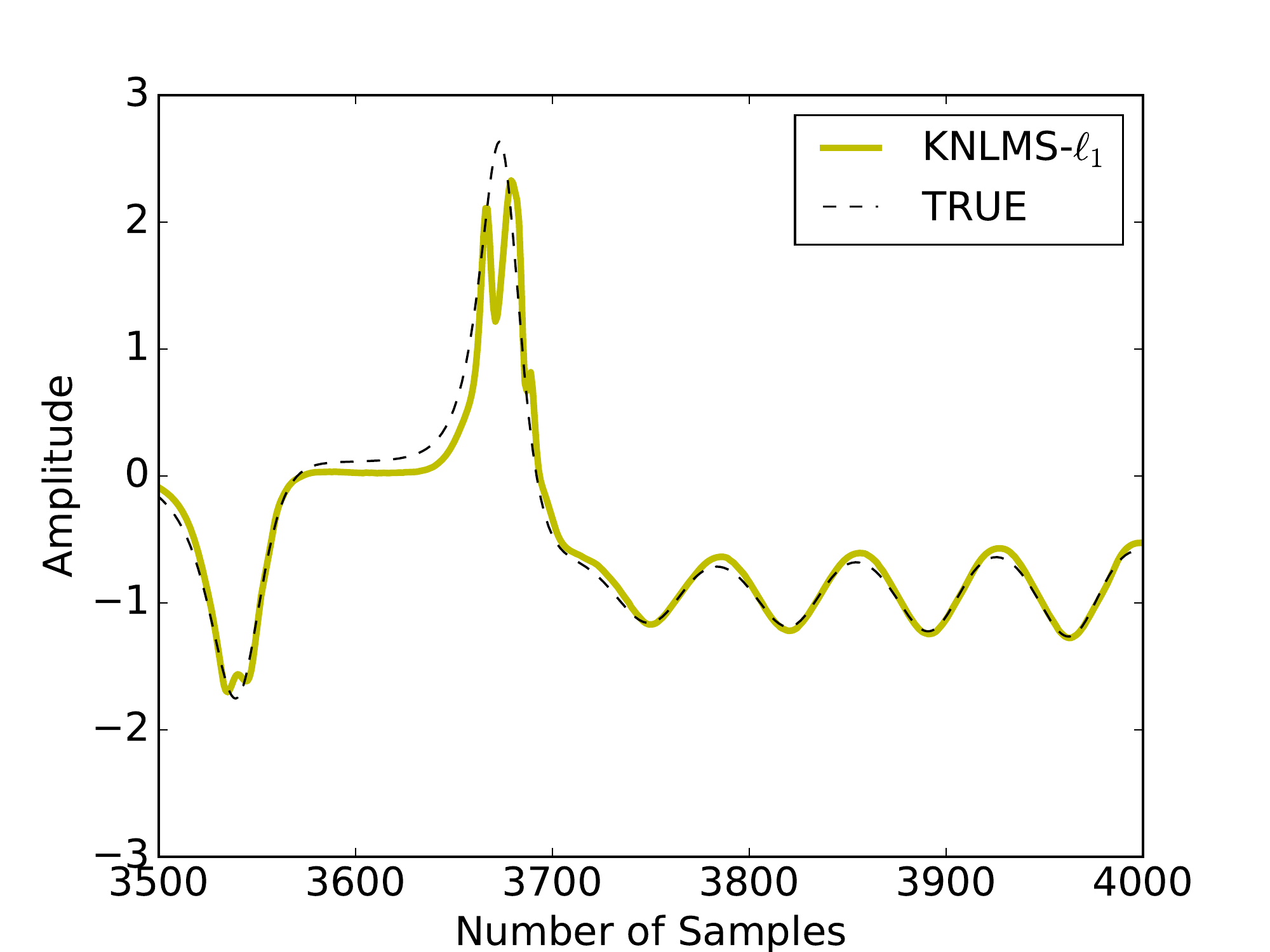}}
	\subfloat[Tracking of NMEG ($L=1$)]{\includegraphics[width=0.5\linewidth]{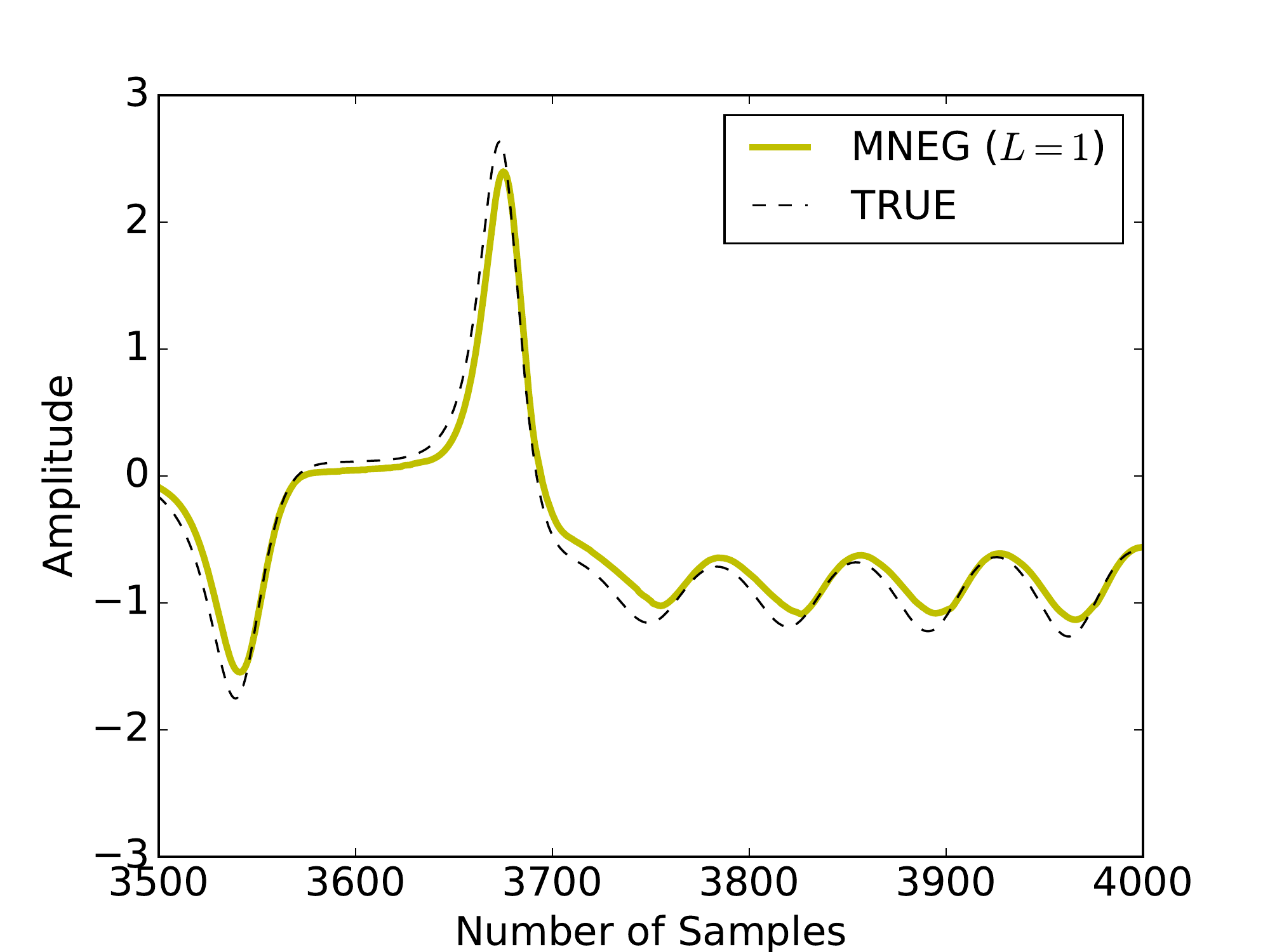}}\\
	%\vspace{-12pt}
	\subfloat[Tracking of MEG]{\includegraphics[width=0.5\linewidth]{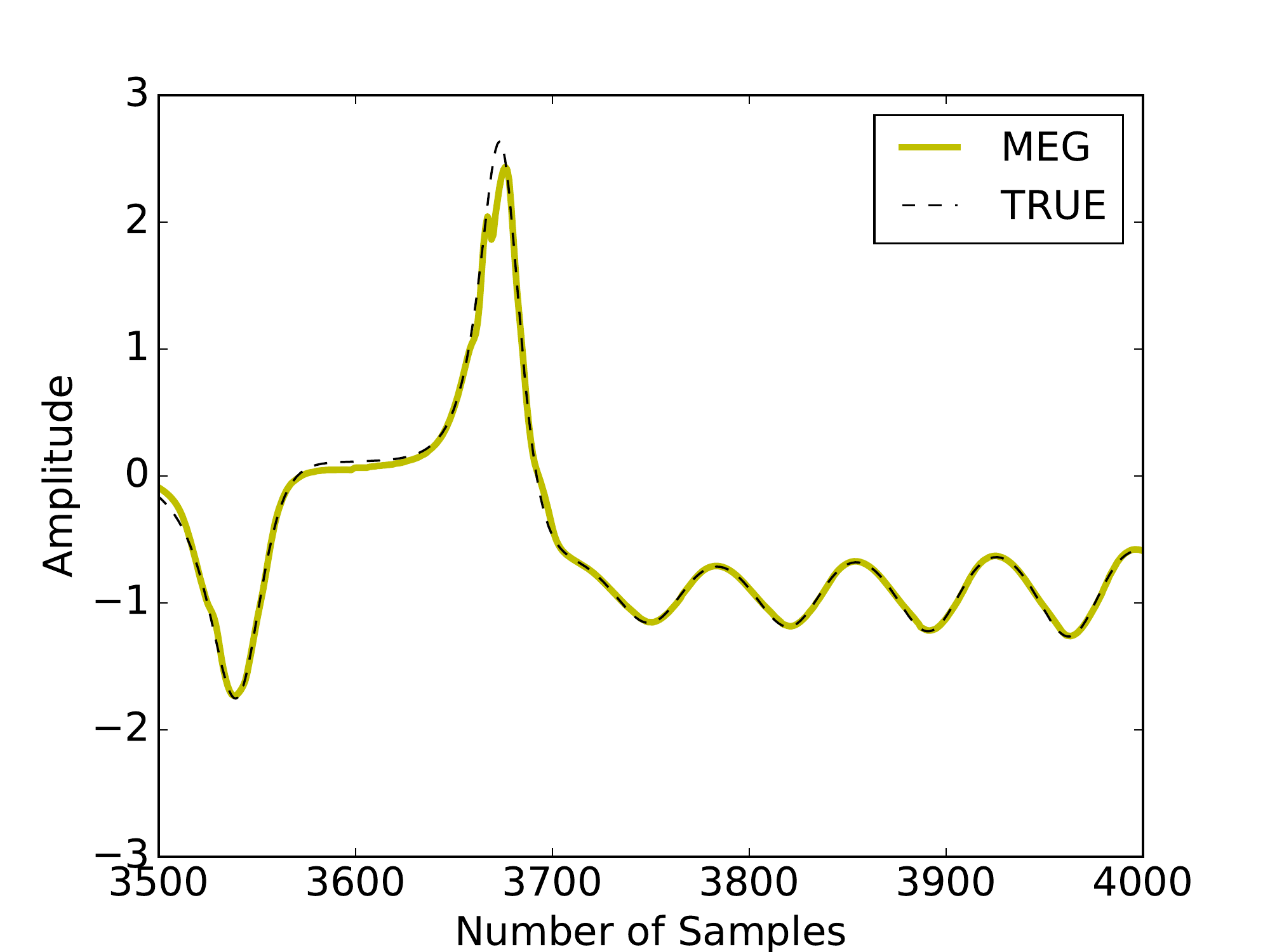}}
	\subfloat[Tracking of NMEG]{\includegraphics[width=0.5\linewidth]{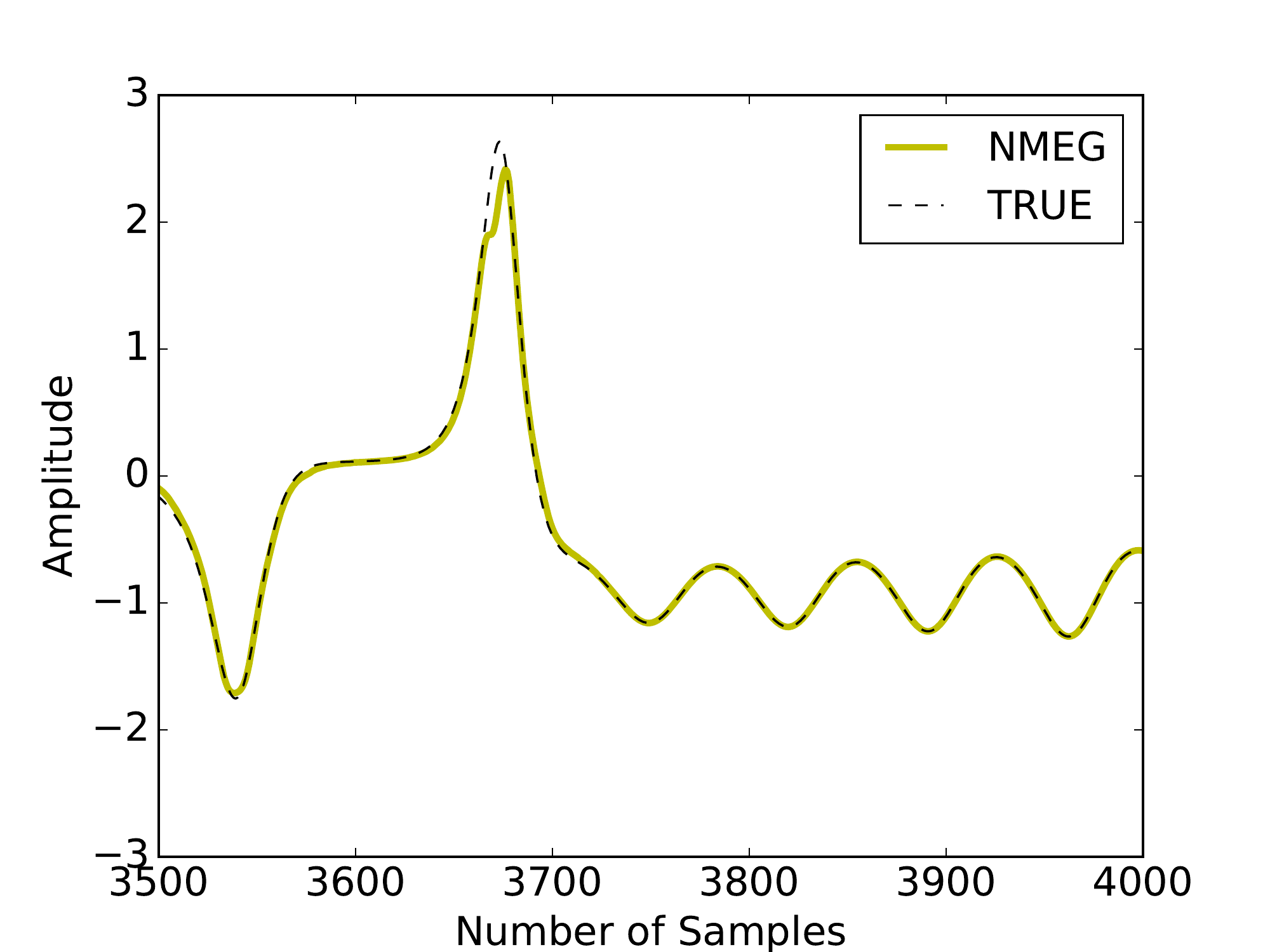}}\\
	\caption{Tracking of the filters in the prediction of the state of a Lorenz chaotic system. (Plots of the last 500 samples in a simulation.}
	\label{fig:lorenz_plot}
	%\vspace{-3pt}
\end{figure*}

\begin{table}[t]
	\centering
	\caption{Parameters in experiment \ref{sec:lorenz}}
	%\vspace{-2pt}
	\begin{tabular}{c|c} \hline
		KNLMS-$\ell_1$&$\mu=0.5,~\rho=0.05,~\zeta=1$\\
		&$\lambda=5.0\times10^{-4},~\beta=0.1$\\\hline
		NMEG ($L=1$)&$\mu=0.5,~\rho=0.05,~\zeta_{\rm init}=1.0,~\lambda=5.0\times10^{-4}$\\
		&$\beta=0.1,~\eta_{c}=0.5,~\eta_{\rm w}=0.1$\\\hline
		MEG&$\mu=0.5,~\rho=0.05,~\bm{Z}_{\rm init}=\bm{I},\lambda=5.0\times10^{-4}$\\
		&$\beta=0.1,~\eta_{c}=0.5,~\eta_{\rm w}=0.1,~L=5$\\\hline
		NMEG&$\mu=0.5,~\rho=0.05,~\bm{Z}_{\rm init}=\bm{I},\lambda=5.0\times10^{-4}$\\
		&$\beta=0.1,~\eta_{c}=0.5,~\eta_{\rm w}=0.1,~L=5$\\\hline
	\end{tabular} \label{tb:lorenz}
	%\vspace{-3pt}
\end{table}
\begin{figure}[t]
	\centering
	\includegraphics[width=\columnwidth,clip]{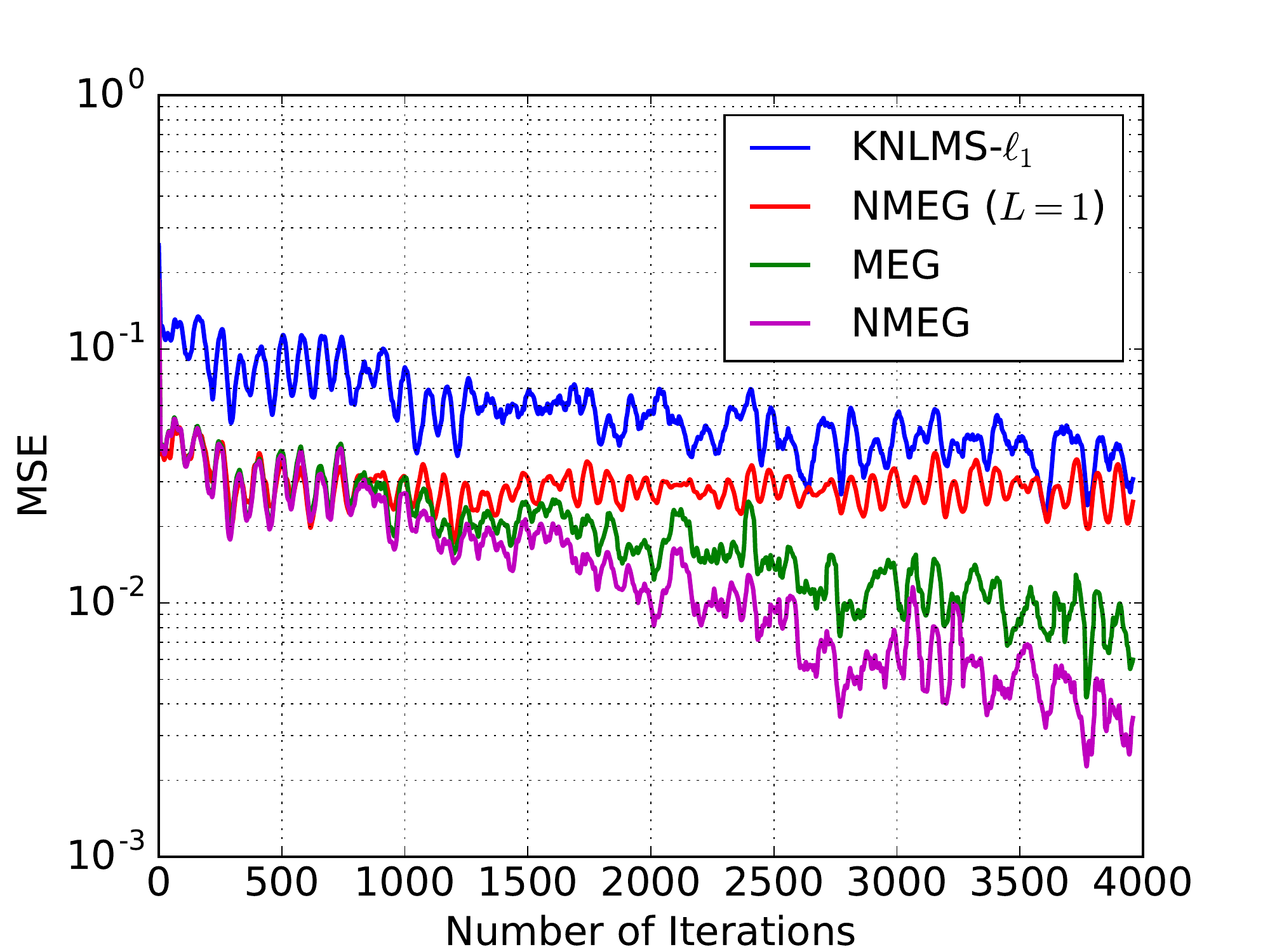}
	%\vspace{-2pt}
	\caption{Convergence curves of filters in experiment \ref{sec:lorenz}. The results are obtained as the average over 50 independent runs with different segments of the signal.}
	\label{fig:lorenz_mse}
	%\vspace{-3pt}
\end{figure}
\begin{figure}[t]
	\centering
	\includegraphics[width=\columnwidth,clip]{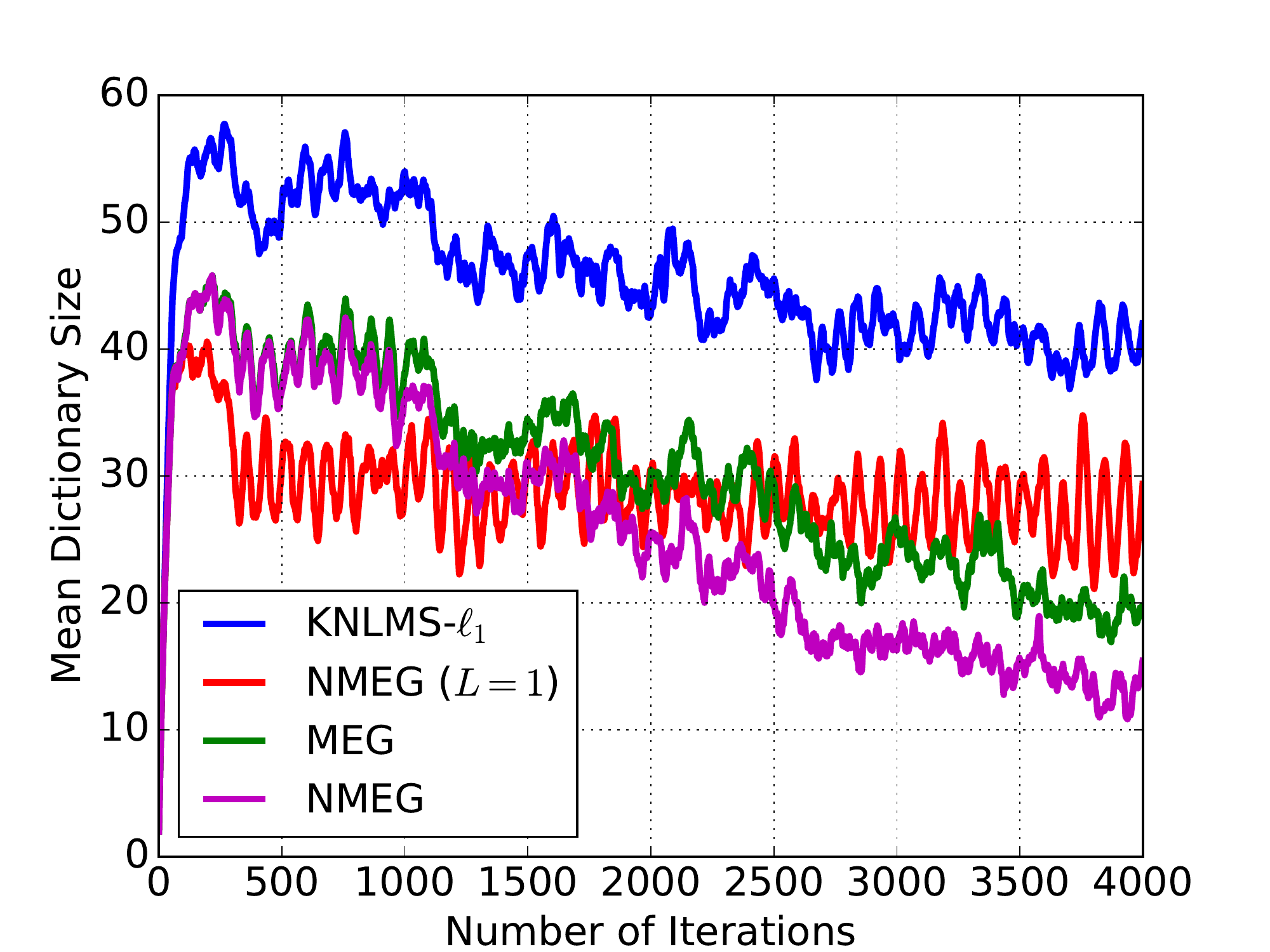}
	%\vspace{-2pt}
	\caption{Dictionary size evolution in experiment \ref{sec:lorenz}. The results are obtained as the average over 50 independent runs with different segments of the signal.}
	\label{fig:lorenz_mds}
	%\vspace{-3pt}
\end{figure}
Consider the Lorenz chaotic system whose states are governed by the differential equations~\cite{liu2009extended}:
\begin{equation}
\begin{cases}
\frac{dx}{dt} =& -\alpha x+yz\\
\frac{dy}{dt} =& -\delta(y-z) \\
\frac{dz}{dt} =& -xy+\gamma y-z,
\end{cases}
\end{equation}
where the parameters are set as $\alpha=8/3$, $\delta=10$, and $\gamma=28$~\cite{chen2012Q}.
The sample data were obtained using first-order (Euler) approximation with step size $0.01$.
The first component (namely $x$) is used in the following for the short-term prediction task.
The signal is normalized to zero-mean and unit variance.
A segment of the processed Lorenz time series is shown in the Figure~\ref{fig:lorenz}.
The problem setting for short-term prediction is as follows:
the previous five points $\bm{u}^{(n)} = [x^{(n-5)}, x^{(n-4)}, \dots, x^{(n-1)}]^{\top}$ are used as the input vector to predict the current value $x^{(n)}$,
which is the desired response.
The MSE is calculated by taking an arithmetic average over $50$ independent realizations with different segments of the signal.
Parameters of filters in this experiment are given in the Table~\ref{tb:lorenz}.
The Figures~\ref{fig:lorenz_mse} and~\ref{fig:lorenz_mds} show the MSE and the mean dictionary size of filters at each iteration, respectively.
Simulation results indicate that the proposed MEG and NMEG exhibit much better performance,
namely, they achieve both much smaller mean dictionary size and much smaller MSE values than the other algorithms used for comparison.
Comparing the MEG algorithm with the NMEG, the NMEG exhibits better performance in terms of both MSE and mean dictionary size although
their parameters are set to the same values.
The Figure~\ref{fig:lorenz_plot} shows the tracking of filters.
It can be seen that the NMEG has higher tracking ability than the NMEG ($L=1$) when a system dynamically changes.
This result supports the validity of the proposed model in the case that the components of the input signals are mutually correlated.
\section{Conclusions}\label{sec:conclusion}
This paper proposed a flexible dictionary learning in the context of generalized Gaussian kernel adaptive filtering, where the kernel parameters are all adaptive and data-driven.
Every input sample or signal has its own precision matrix and center vector, which are updated at each iteration based on the proposed least-square type rules to minimize the estimation error.
In particular, we proposed a novel update rule for precision matrices,
which allows one to update each precision matrix stably due to an effective normalization.
Together with the $\ell_1$ regularized least squares, the overall kernel adaptive filtering algorithms can avoid overfitting and the monotonic growth of a dictionary.
Numerical examples showed that the proposed method exhibits higher performance in terms of the MSE and the size of dictionary in the prediction of nonlinear systems.
%
% biography section
\balance
\bibliographystyle{IEEEtran}
\bibliography{IEEEabrv,wada}
\end{document}